\documentclass[letterpaper]{article} 
\usepackage{aaai2026}  
\usepackage{times}  
\usepackage{helvet}  
\usepackage{courier}  
\usepackage[hyphens]{url}  
\usepackage{graphicx} 
\usepackage{natbib}  
\usepackage{caption} 
\usepackage{algorithm}
\usepackage{cite}
\usepackage{amsmath,amssymb,amsfonts,amsthm}
\usepackage{algpseudocode}
\usepackage{graphicx}
\usepackage{subcaption}
\usepackage{textcomp}
\usepackage{mathtools}
\usepackage{xcolor}
\usepackage{multirow}
\usepackage{booktabs}
\usepackage{fontawesome5}
\newtheorem{theorem}{Theorem}[section]
\newtheorem{proposition}[theorem]{Proposition}
\newtheorem{lemma}[theorem]{Lemma}

\usepackage{newfloat}
\usepackage{listings}
\DeclareCaptionStyle{ruled}{labelfont=normalfont,labelsep=colon,strut=off} 
\floatstyle{ruled}
\newfloat{listing}{tb}{lst}{}
\floatname{listing}{Listing}
\title{Full-Atom Peptide Design via Riemannian–Euclidean Bayesian Flow Networks}
\author {
    Hao Qian\textsuperscript{\rm 1},
    Shikui Tu\textsuperscript{\rm 1 \faEnvelope[regular]},
    Lei Xu\textsuperscript{\rm 1, 2 \faEnvelope[regular]}
}
\affiliations {
    \textsuperscript{\rm 1} School of Computer Science, Shanghai Jiao Tong University\\
    \textsuperscript{\rm 2} Guangdong Laboratory of Artificial Intelligence and Digital Economy (SZ), Guangdong, China\\
    tushikui@sjtu.edu.cn, leixu@sjtu.edu.cn
}

\begin{document}

\maketitle

\begin{abstract}
Diffusion and flow matching models have recently emerged as promising approaches for peptide binder design. Despite their progress, these models still face two major challenges. First, categorical sampling of discrete residue types collapses their continuous parameters into one-hot assignments, while continuous variables (e.g., atom positions) evolve smoothly throughout the generation process. This mismatch disrupts the update dynamics and results in suboptimal performance. Second, current models assume unimodal distributions for side-chain torsion angles, which conflicts with the inherently multimodal nature of side-chain rotameric states and limits prediction accuracy. To address these limitations, we introduce PepBFN, the first Bayesian flow network for full-atom peptide design that directly models parameter distributions in fully continuous space. Specifically, PepBFN models discrete residue types by learning their continuous parameter distributions, enabling joint and smooth Bayesian updates with other continuous structural parameters. It further employs a novel Gaussian mixture–based Bayesian flow to capture the multimodal side-chain rotameric states and a Matrix Fisher–based Riemannian flow to directly model residue orientations on the $\mathrm{SO}(3)$ manifold. Together, these parameter distributions are progressively refined via Bayesian updates, yielding smooth and coherent peptide generation. Experiments on side-chain packing, reverse folding, and binder design tasks demonstrate the strong potential of PepBFN in computational peptide design.
\end{abstract}

\begin{links}
    \link{Code}{https://github.com/CMACH508/PepBFN}
\end{links}

\section{Introduction}

Peptides, short chains of amino acids, can be engineered to specifically bind to protein targets. This specific binding modulates protein activity and can influence key biological processes \cite{craik2013future, henninot2018pep, wang2022therapeutic}. Compared to macromolecular biologics, such as proteins and RNAs, peptides exhibit lower immunogenicity and are less costly to produce \cite{giordano2014neuroactive, fosgerau2015peptide, davda2019immunogenicity}. Relative to small molecule drugs, peptides typically offer greater safety and enhanced tolerance \cite{di2015strategic,muttenthaler2021trends}. Hence, peptides serve as an effective bridge between small molecules and biomacromolecules.

Traditional peptide design starts by analyzing the known crystal structures to understand their primary and secondary structures. Subsequently, methods such as alanine scanning (Alascan) \cite{lefevre1997alanine} and the utilization of small, focused libraries \cite{quartararo2020ultra} help to systematically develop the structure-activity relationship (SAR) \cite{fosgerau2015peptide}. However, the efficiency of these methods is significantly reduced by the combinatorial explosion of amino acid sequences. As experimental limitations persist, there is an increasing demand for computational approaches that enhance in silico peptide binder design \cite{bryant2022evobind,swanson2022tertiary,cao2022design,bhat2023novo,chen2024pepmlm}.

Recently, deep generative models, such as auto-regressive models, diffusion models \cite{song2020score, song2020denoising,dhariwal2021diffusion}, and flow matching models \cite{lipman2022flow,liu2022flow}, have demonstrated their potential in peptide binder design \cite{li2024pepflow, lin2024ppflow,kong2024pepglad,wang2024target}. PepHAR~\cite{li2024hotspot}, an auto-regressive model, efficiently generates peptide residues in a sequential manner within 3D space, guided by learned anchor hotspots. However, since it predicts residues step-by-step without considering the overall peptide conformation, it tends to produce severe steric clashes between the generated peptide and the target. Diffusion and flow matching models, such as PepFlow~\cite{li2024pepflow} and PepGLAD~\cite{kong2024pepglad}, alleviate steric clashes through non-autoregressive generation. 

Despite these advances, there still remain two limitations in existing models. Firstly, discrete residue types are sampled categorically, which collapses their continuous parameters into one-hot assignments and discards distributional information, while continuous structural variables are sampled smoothly. This mismatch leads to inconsistent samplings, causing sequence fluctuations and slow convergence, as also observed in previous studies on structure-based molecule design~\cite{peng2023moldiff, song2023equivariant,qu2024molcraft}. Secondly, existing approaches typically assume a unimodal distribution for side-chain angles, whereas their true distributions are inherently multimodal due to diverse rotameric states. This assumption negatively impacts the accuracy of side-chain prediction (see Sec.~\ref{sec:scpack}).

To overcome these limitations, we introduce PepBFN, a Riemannian–Euclidean Bayesian flow network that jointly models sequences and structures in fully continuous parameter space. Bayesian Flow Networks~\cite{graves2023bayesian} (BFNs) iteratively refine continuous parameter distributions via Bayesian inference, leading to smooth updates that resolve the mismatch problem. Moreover, unlike prior generative approaches that utilize unimodal distributions for modeling side-chain angles, we propose a novel Gaussian mixture-based Bayesian flow inference process. This design enables an effective representation of diverse rotameric states throughout the generation, leading to more accurate side-chain prediction. In addition, to model residue orientations, we introduce a new Riemannian Bayesian flow based on the Matrix Fisher distribution~\cite{downs1972orientation,khatri1977mises}. As an exponential family distribution on the $\mathrm{SO}(3)$ manifold, Matrix Fisher enables tractable Bayesian updates, allowing residue orientations to be seamlessly integrated into the Bayesian flow framework.

\vspace{1em}
\noindent \textbf{Our main contributions are as follows:}
\begin{itemize}
    \item We propose PepBFN, the first Bayesian Flow Network for full-atom peptide binder design, which jointly models four key modalities, i.e., discrete residue types, continuous residue orientations, centroids, and side-chain torsions, within fully continuous parameter space.
    \item PepBFN incorporates a novel Gaussian mixture–based Bayesian flow for accurately capturing multimodal side-chain conformations, and a Matrix Fisher–based Riemannian flow for residue rotations, enabling tractable Bayesian updates on the $\mathrm{SO}(3)$ manifold.
    \item PepBFN achieves state-of-the-art performance on multiple benchmarks, including de novo peptide binder design, side-chain prediction, and reverse folding tasks, demonstrating its effectiveness as a unified framework for computational peptide design.
\end{itemize}

\section{Related Works}
\subsection{Bayesian Flow Networks}
Bayesian Flow Networks (BFNs)~\cite{graves2023bayesian} are a recently proposed generative modeling framework that integrates Bayesian inference with neural networks. BFNs evolve parameter distributions through iterative Bayesian updates guided by a noise scheduler, rather than perturbing data directly as in diffusion models with predefined forward processes. This parameter-space formulation naturally accommodates both discrete and continuous variables, offers a smooth and differentiable generative trajectory. BFNs demonstrate competitive performance across diverse domains, showing notable advantages in discrete settings such as protein language modeling~\cite{atkinson2025protein} and molecule design~\cite{song2024unified,qu2024molcraft}.

\subsection{Generative Models for Protein and Peptide Binder Design}

Generative models have recently emerged as powerful tools for protein and peptide binder design, providing enhanced flexibility compared to traditional template- or energy-based methods. Early approaches predominantly focused on employing protein language models for sequence design~\cite{madani2020progen,ferruz2022controllable}. 
Recent studies have focused on the joint design of sequence and structure. Specifically for peptide design, models such as PepFlow~\cite{li2024pepflow}, SurfFlow~\cite{surfflow} and PPFlow~\cite{lin2024ppflow} leverage multi-modal flow matching to perform full-atom peptide generation conditioned on target protein structures. Additionally, diffusion-based models like PepGLAD~\cite{kong2024pepglad} and DiffPepBuilder~\cite{wang2024target} explore joint diffusion processes to design peptide sequences and structures. Autoregressive approaches such as PepHAR~\cite{li2024hotspot} sequentially generate peptide residues in 3D space, guided by learned anchor hotspots. Collectively, these generative models have advanced protein and peptide binder design by enabling more flexible and scalable computational pipelines.

\section{Preliminaries}
\subsection{Problem Formulation}
Protein–peptide complexes are denoted by $ \{\mathcal{P}, \mathcal{G}\}$, where $\mathcal{P}$ and $\mathcal{G}$ represent the protein and peptide respectively. Each component is described as a sequence of local residue frames. For the $i$-th residue, we represent its geometry by the position of residue centroid (i.e., C$_\alpha$ position) $\mathbf{x}^{(i)} \in \mathbb{R}^3$ and an orientation matrix $\mathbf{o}^{(i)} \in \mathrm{SO}(3)$, consistent with the representation used in AlphaFold 3~\cite{abramson2024af3}. The torsional angle set $\mathbf{\chi}^{(i)} = \{{\psi^{(i)}, \chi^{(i)}_1, \ldots, \chi^{(i)}_4}\}$ includes the backbone angle $\psi^{(i)}$, which determines the position of the backbone oxygen atom, and up to four side-chain $\chi_{1-4}^{(i)}$ angles. The amino acid type is encoded as a one-hot vector $\mathbf{c}^{(i)} \in \mathbb{R}^{20}$. Thus, a protein or peptide with $N$ residues can be expressed as $\{\mathcal{R}^{(i)}\}_{i=1}^{N}$, where each residue is defined by $\mathcal{R}^{(i)} = \{\mathbf{x}^{(i)}, \mathbf{o}^{(i)}, \mathbf{\chi}^{(i)}, \mathbf{c}^{(i)} \}$.

Given a target protein pocket $\mathcal{P}$, PepBFN designs new peptides $\mathcal{G}$ that bind effectively to the target protein by jointly modeling their sequences and 3D structures.

\subsection{Peptide Design via BFNs}
We define $\boldsymbol{\theta}$ as the parameters of the input data distribution $p_I(\mathcal{G}\mid\boldsymbol{\theta})$. The peptide generation process is formulated as a Bayesian communication between a sender and a receiver. At each time step $t_i$, the sender generates a noisy peptide $\mathbf{y}_i$ by perturbing the clean peptide $\mathcal{G}$ with a known noise factor $\alpha_i$, resulting in the \textit{sender distribution}:
\[
p_S(\mathbf{y}_i \mid \mathcal{G}; \alpha_i),
\] which resembles the idea of forward process in diffusion models. The receiver acts as the decoder, conditioned on the protein context $\mathcal{P}$ and previously inferred parameters $\boldsymbol{\theta}_{i-1}$, aiming to reconstruct the clean peptide $\hat{\mathcal{G}}$, yielding the \textit{output distribution}:
\[
    p_O(\hat{\mathcal{G}} \mid \boldsymbol{\theta}_{i-1}, \mathcal{P}; t_i) = \Phi(\boldsymbol{\theta}_{i-1}, \mathcal{P}, t_i).
\] Here, $\Phi$ denotes a neural network conditioned on the protein target $\mathcal{P}$, the previous-step parameters $\boldsymbol{\theta}_{i-1}$, and the current time step $t_i$. 
Given the known noise factor $\alpha_i$, the receiver is able to generate the noisy peptide $\mathbf{y}_i$ by injecting noise into its estimate $\hat{\mathcal{G}}$, thereby forming the \textit{receiver distribution}:
\[
p_R(\mathbf{y}_i \mid \boldsymbol{\theta}_{i-1}, \mathcal{P}; t_i) = \mathbb{E}_{\hat{\mathcal{G}} \sim p_O} \left[ p_S(\mathbf{y}_i \mid \hat{\mathcal{G}}; \alpha_i) \right].
\]

During generation, the receiver applies Bayesian inference to iteratively refine the peptide parameters $\boldsymbol{\theta}$ in closed form. Specifically, the \textit{Bayesian update distribution} $p_U$ is derived from the Bayesian update function $h$ as:
\begin{equation}
\resizebox{0.9\linewidth}{!}{$
p_U(\boldsymbol{\theta}_i \mid \boldsymbol{\theta}_{i-1}, \mathcal{G}, \mathcal{P}; \alpha_i) =
\mathbb{E}_{\mathbf{y}_i \sim p_S} \Big[
\delta\big(\boldsymbol{\theta}_i - h(\boldsymbol{\theta}_{i-1}, \mathbf{y}_i, \alpha_i)\big)
\Big].
$}
\label{eq:bayesian_update_distribution}
\end{equation}
Here, $\delta(\cdot)$ denotes the Dirac delta distribution, and $h$ maps the previous parameters $\boldsymbol{\theta}_{i-1}$ and noisy observation $\mathbf{y}_i$ to the updated parameters $\boldsymbol{\theta}_i$.

In addition, the \textit{Bayesian flow distribution} $p_F$ can be achieved to support simulation-free training as follows:
\begin{align}
p_F(\boldsymbol{\theta}_i \mid \mathcal{G}, \mathcal{P}; t_i) 
&= \mathbb{E}_{\boldsymbol{\theta}_{1:i-1} \sim p_U} 
\, p_U(\boldsymbol{\theta}_i \mid \boldsymbol{\theta}_{i-1}, \mathcal{G}, \mathcal{P}; \alpha_i) \nonumber \\
&= p_U(\boldsymbol{\theta}_i \mid \boldsymbol{\theta}_0, \mathcal{G}, \mathcal{P}; \beta(t_i)),
\label{eq:bayesian_flow}
\end{align}
where $\beta(t_i) = \sum_{j=1}^i \alpha_j$ represents the accumulated noise scheduler based on the additive property of the noise factors~\cite{graves2023bayesian}.
The model is trained by minimizing the expected KL divergence between the sender and receiver distributions over $n$ time steps:
\begin{equation} \label{eq:loss}
L_n(\mathcal{G}, \mathcal{P}) = n \, \mathbb{E}_{i \sim U(1, n),\, \boldsymbol{\theta}_{i-1} \sim p_F} \, D_{\mathrm{KL}}(p_S \, \| \, p_R).
\end{equation}

We summarize the key similarities and differences among different generative models in Table~\ref{tab_app:compare} in the Appendix.

\textbf{Difficulties for Peptide Design under the BFN Framework}
A key theoretical challenge in applying BFNs to peptide design is the requirement for conjugate prior–posterior distributions. However, the commonly used distributions for side-chain angles (e.g., unimodal wrapped Gaussians~\cite{zhang2023diffpack}) and residue orientations (e.g., Brownian motion on $\mathrm{SO}(3)$~\cite{leach2022denoising}) violate this requirement, making principled training and inference difficult to implement in practice. Moreover, as we mentioned before, unimodal Gaussian distribution can not accurately model side-chain rotameric states. In this paper, we address these issues by devising two new BFNs for side-chain angles and residue orientations, respectively, and they both support conjugate-aligned formulations tailored to peptide design.

\section{Methods}

\begin{figure}
    \centering
    \includegraphics[width=\linewidth]{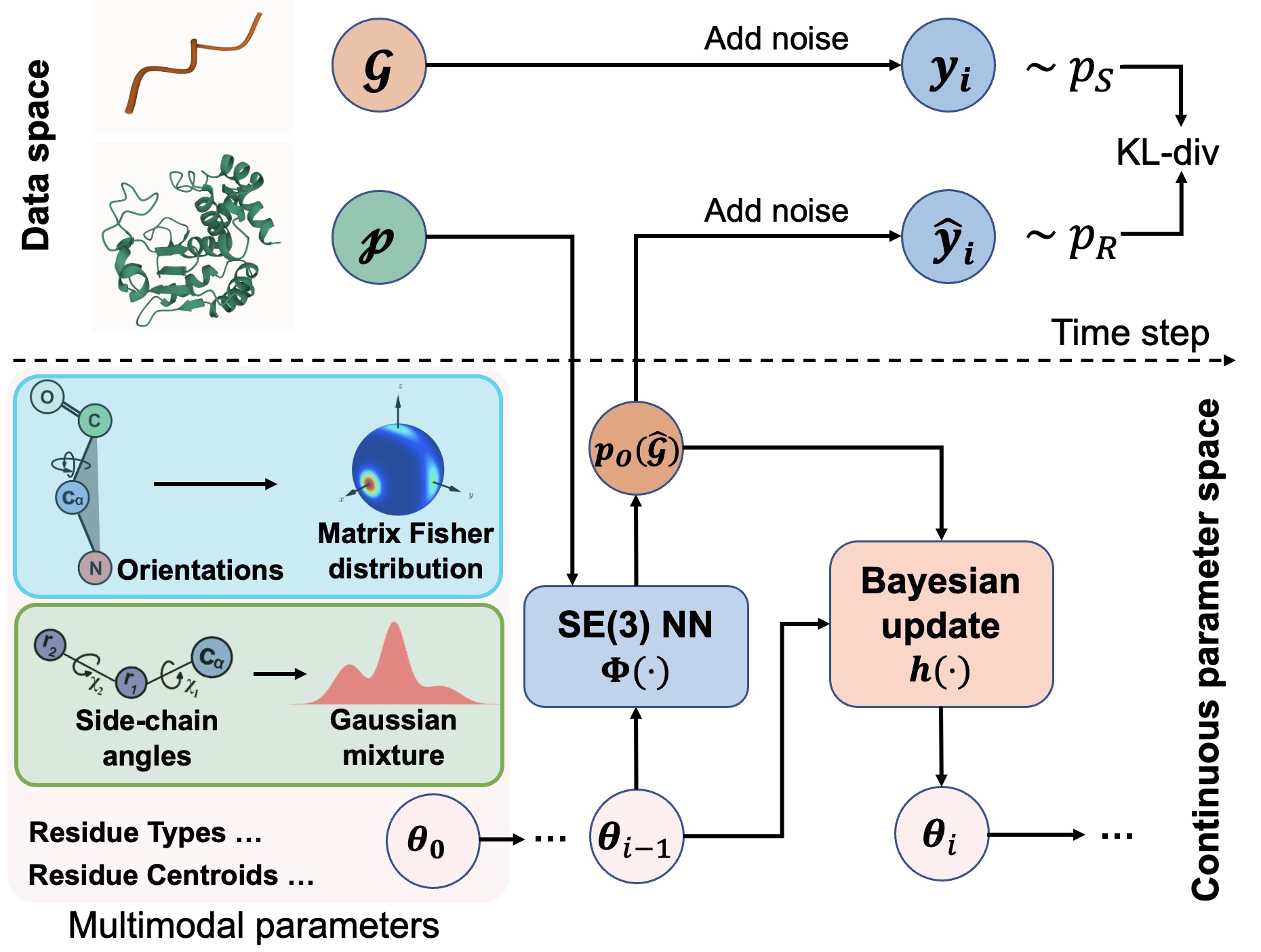}
    \vspace{-1.5em}
    \caption{The overview of PepBFN.}
    \label{fig:pipeline}
    \vspace{-1.5em}
\end{figure}

\subsection{The Overview of PepBFN}
As shown in Fig.~\ref{fig:pipeline}, we construct PepBFN to jointly capture four key modalities involved in peptide design. Specifically, for residue orientations, we develop a Matrix Fisher-based Riemannian BFN, allowing for direct Bayesian update on the $\mathrm{SO}(3)$ manifold. For torsional angles, we propose a novel Gaussian mixture-based BFN on the toric manifold, enabling accurate modeling of multimodal angular distributions. Residue centroids are modeled based on unimodal Gaussian distributions, and residue types are modeled with categorical distributions, inspired by the original BFN paper~\cite{graves2023bayesian}. Together, these four components form a unified BFN framework for peptide binder design. 

During generation, the SE(3) neural network takes the protein context $\mathcal{P}$ and previous parameters 
$\boldsymbol{\theta}_{i-1}$ as inputs, and predicts the denoised peptide $\hat{\mathcal{G}}$, which serves as parameters of the receiver distribution $p_R$. Samples drawn from \(p_R\) are combined with the previous parameters $\boldsymbol{\theta}_{i-1}$ through the Bayesian update operator \(h(\cdot)\), producing the posterior parameters \(\boldsymbol{\theta}_i\), which are then propagated to the next time step. 
The entire generative process is carried out in fully continuous parameter space, ensuring smooth and consistent parameter updates. To the best of our knowledge, PepBFN is the first Bayesian Flow Network specifically designed for the full-atom peptide design.

\subsection{Gaussian Mixture-based BFN for Angles}

\begin{figure}
    \centering
    \includegraphics[width=0.8\linewidth]{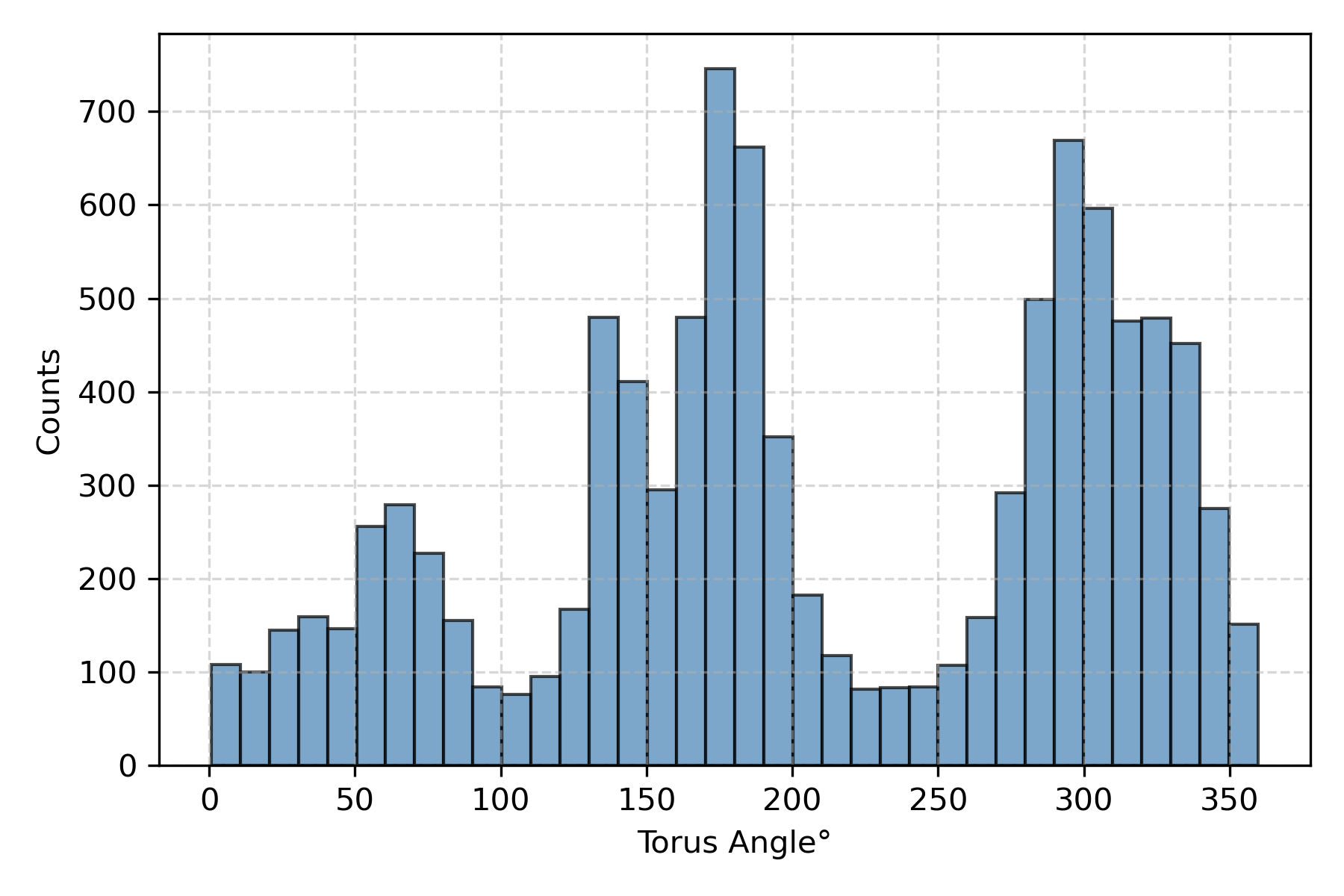}
    \vspace{-1em}
    \caption{Distribution of peptide torus angles.}
    \label{fig:torus_angles}
    \vspace{-1em}
\end{figure}

Side-chain $\chi$ angles in proteins are often modeled using a single wrapped Gaussian distribution to account for rotational periodicity~\cite{zhang2023diffpack}. However, such unimodal approximations fail to capture their true conformational heterogeneity. Our analysis on prior distribution  (see Fig.~\ref{fig:torus_angles}) shows that $\chi$ angles cluster around three distinct rotameric states corresponding to \textit{gauche}$^+$ , \textit{trans}, and \textit{gauche}$^-$\footnote{
Here, \textit{gauche}$^-$ and \textit{gauche}$^+$ denote staggered rotameric states where the torsion angle \(\chi\) is near \(300^\circ\) (or equivalently \(-60^\circ\)) and \(60^\circ\), respectively. The \textit{trans} state corresponds to another low-energy staggered conformation near \(180^\circ\), commonly observed in rotamer libraries.
}. To accurately capture the multimodal nature of side-chain torsion angles, we employ a novel Gaussian mixture-based BFN, which effectively learns multiple rotameric states throughout the time steps.

The input distribution is given by
\begin{align}\label{eq:angle_input}
    \boldsymbol{\theta}^{ang} 
&\stackrel{\mathrm{def}}{=} 
\big\{ \mu^k, \rho^k, \pi^k \big\}_{k=1}^K, \nonumber \\ 
    p_I(\chi\mid \boldsymbol{\theta}^{ang}) &\stackrel{\text{def}}{=} \sum_{k=1}^{K} \pi^k \, \mathcal{N}\big(\chi \mid \mu^k, (\rho^k)^{-1}\big),
\end{align}
where $\pi^k \geq 0$, $\sum_{k=1}^{K} \pi^k = 1$ are the mixture weights, and $\rho^k$ is the known precision for k-th Gaussian. The sender distribution is defined as a single Gaussian:
\begin{equation}\label{eq:ang_sender}
    p_S(y_i^{ang}\mid \chi) = \mathcal{N}(y_i^{ang}\mid \chi, \mathcal{P}, \alpha_i; t_i),
\end{equation}
where $\alpha_i$ is a noise factor at $t_i$ time step.

\begin{lemma}[Conjugacy of a Gaussian Mixture Prior with a Gaussian Likelihood]\label{lemma:gmm}
Let the prior \(p(x)\) be a Gaussian mixture distribution and the likelihood \(p(y \mid x)\) be a single Gaussian distribution. By Bayes’ rule, the posterior \(p(x \mid y)\) retains the Gaussian mixture form.
\end{lemma}

\begin{proposition}[Bayesian Flow for Gaussian Mixture Distribution]\label{prop:bfn4gmm}
Given the input distribution defined in Eq.~\ref{eq:angle_input} and the sender distribution defined in Eq.~\ref{eq:ang_sender}, 
the \textit{Bayesian flow distribution} $p_F$ for the mean parameter $\mu_i^k$ of the $k$-th component at time step $t_i$ is
\begingroup\small
\begin{equation}
    p_F(\mu^k_i \mid \chi, \mathcal{P}; t_i) = 
    \mathcal{N}\!\left(
    \mu_i^k \,\middle|\, 
    \frac{\beta(t_i)\chi + \mu_0^k\rho_0^k}{\rho_i^k}, 
    \frac{\beta(t_i)}{(\rho_i^k)^2}
    \right),
\end{equation}
\endgroup
where the $k$-th precision of input distribution $p_I(\cdot)$ at $t_i$ time step is $\rho_i^k = \rho_0^k + \beta(t_i)$ and the precision scheduler is $\beta(t_i) = \sum_{m=1}^i \alpha_m$. 
The \textit{Bayesian flow distribution} of the mixture weights $\pi_i^k$ at time step $t_i$ does not admit a closed-form expression and requires numerical simulation.
\end{proposition}

Proposition~\ref{prop:bfn4gmm} provides an analytical expression for the mean parameters $\mu_i$ and a simulation-based procedure for the weight parameters $\pi_i$ in the Bayesian flow distribution $p_F$. We next design the precision scheduler $\beta(\cdot)$ to ensure that the expected entropy of the input distribution $p_I(\cdot)$ under $p_F$ decreases linearly over time as suggested in~\cite{graves2023bayesian}. Specifically, the expected entropy is defined as:
\begin{equation} \label{eq:expected_rntropy}
H(t_i) \stackrel{\mathrm{def}}{=} \mathbb{E}_{p_F(\boldsymbol{\theta}^{ang}_i \mid \chi,\mathcal{P}; t_i)} \big[ H(p_I(\cdot \mid \boldsymbol{\theta}^{ang}_i)) \big].
\end{equation}

\begin{proposition}[Time-Dependent Linear Decrease of the Expected Entropy Upper Bound]\label{prop:upper_bound}
    Given the expected entropy $H(t_i)$ defined in Eq.~\ref{eq:expected_rntropy} and the Gaussian likelihood precision parameter $\alpha_i =\rho_0^{1-i/n} \cdot \rho_1^{i/n}\big[1-(\frac{\rho_0}{\rho_1})^{1/n}   \big]$ at the time step $t_i$, the upper bound of the expected entropy $H(t_i)$ decreases linearly with the time step $t_i$.
\end{proposition}

Given the Gaussian likelihood precision parameter $\alpha_i$ defined in Proposition~\ref{prop:upper_bound}, the receiver distribution can be achieved:
\[
p_R(y^{ang}_i \mid \boldsymbol{\theta}^{ang}_{i-1}, \mathcal{P}; t_i) = \mathbb{E}_{\hat{\chi} \sim p_O} \left[ p_S(y^{ang}_i \mid \hat{\chi}; \alpha_i) \right],
\]
\[
\text{where} \quad p_O(\hat{\chi} \mid \boldsymbol{\theta}_{i-1}^{ang}, \mathcal{P}; t_i) = \Phi^{ang}(\boldsymbol{\theta}_{i-1}^{ang}, \mathcal{P}, t_i).
\] Here, $\Phi^{ang}$ denotes a neural network conditioned on protein target $\mathcal{P}$, parameters $\boldsymbol{\theta}_{i-1}^{ang}$ and time step $t_i$. Substituting into Eq.~\ref{eq:loss}, the loss function becomes:
\begin{align}
&L_n^{ang}(\mathbf{\chi}, \mathcal{P}) = n \, \mathbb{E}_{i \sim U(1, n),\, \boldsymbol{\theta}_{i-1}^{ang} \sim p_F} \, D_{\mathrm{KL}}(p_S \, \| \, p_R) \nonumber \\
&= \frac{n}{2} \, \mathbb{E}_{i \sim U(1, n),\, \boldsymbol{\theta}_{i-1}^{ang} \sim p_F} \, \alpha_i \| \chi - \hat{\chi} \|^2.
\end{align}
  
The integration of Gaussian mixture distributions into the BFN framework allows principled and stable modeling of multimodal torsion angles, well-suited for Bayesian inference. Please refer to Algorithm~\ref{alg:sampling} in the Appendix for inference details.


\subsection{Matrix Fisher-based BFN for Orientations}
We represent the orientations of $N$ residues in a peptide as $\mathbf{O} = \{\mathbf{o}^{(i)}\}_{i=1}^N  \sim \mathcal{M}(\mathbf{O}; \boldsymbol{\theta}^{ori})$, where i-th residue orientation is denoted as a rotation matrix $\mathbf{o}^{(i)}$ and $\boldsymbol{\theta}^{ori}$ is the parameter of the Matrix Fisher distribution. Specifically, we define the input distribution as $p_I(\mathbf{O}\mid \boldsymbol{\theta}^{ori}) = \mathcal{M}(\mathbf{O};\boldsymbol{\theta}^{ori}) = \frac{1}{c(\boldsymbol{\theta}^{ori})} \exp\left( \mathrm{tr}((\boldsymbol{\theta}^{ori})^\top \mathbf{O}) \right),$ where $c(\boldsymbol{\theta}^{ori})$ is a normalizing constant and matrix $\boldsymbol{\theta}^{ori} \in \mathbb{R}^{3\times3}$ (see Sec.~\ref{app:intro_matrix_fisher} in the Appendix for details).

\begin{lemma}[Conjugacy of Matrix Fisher Distributions]
Let the prior distribution over rotation matrices $\mathbf{O} \in \mathrm{SO}(3)$ be a Matrix Fisher distribution given by
\[
p(\mathbf{O}) \propto \exp\left(\mathrm{tr}\left(\boldsymbol{\theta}_a^\top \mathbf{O}\right)\right),
\]
and the likelihood of an observation $\mathbf{Y}^{ori} \in \mathrm{SO}(3)$ be
\[
p(\mathbf{Y}^{ori} \mid \mathbf{O}) \propto \exp\left(\mathrm{tr}\left((\mathbf{O\Lambda})^\top \mathbf{Y}^{ori}\right)\right),
\]
where $\boldsymbol{\Lambda}$ denotes a diagonal matrix with identical diagonal entries.
Then, the posterior distribution is also a Matrix Fisher distribution $\mathcal{M}(\mathbf{O} \mid \mathbf{Y}^{ori};\boldsymbol{\theta}_b)$:
\[
p(\mathbf{O} \mid \mathbf{Y}^{ori}) \propto \exp\left(\mathrm{tr}\left((\boldsymbol{\theta}_a + \mathbf{Y}^{ori}\mathbf{\Lambda})^\top \mathbf{O}\right)\right),
\]
where
\begin{equation}
\boldsymbol{\theta}_b = \boldsymbol{\theta}_a + \mathbf{Y}^{ori}\mathbf{\Lambda}.
\label{eq:bayesian_update_func}
\end{equation}
\label{proposition:fisher_conjugacy}
\end{lemma}

Solving the Bayesian update distribution $p_U(\boldsymbol{\theta}^{ori}_i\mid \boldsymbol{\theta}^{ori}_{i-1}, \mathcal{G}, \mathcal{P}; t_i)$ in closed form is non-trivial because the Matrix Fisher distribution family is not closed under operations in Bayesian update function $h(\cdot)$ (see Eq.~\ref{eq:bayesian_update_func}). Here, we introduce an auxiliary variable $\mathbf{T}_i$, which resolves this difficulty and yields a new Matrix Fisher distribution that is both computationally tractable (see Proposition~\ref{thm:fisher}) and straightforward to sample from (see Sec.~\ref{app:fast_sample_matrix} in the Appendix).

\begin{proposition} [Bayesian Flow for Matrix Fisher Distribution] \label{thm:fisher}
    Assume that the input distribution over residue orientations in dataset follows the Matrix Fisher distribution on $\mathrm{SO}(3)$, i.e., $p_I(\mathbf{O}\mid \boldsymbol{\theta}^{ori}) \stackrel{\text{def}}{=} \mathcal{M}(\mathbf{O}; \boldsymbol{\theta}^{ori})$. At time step $t_i$, we define the \textit{sender distribution} as
    $
    p_S(\mathbf{Y}_i^{ori} | \mathbf{O}; \boldsymbol{\Lambda}_i) = \mathcal{M}(\mathbf{Y}_i^{ori}; \mathbf{O} \boldsymbol{\Lambda}_i).
    $
    We introduce an auxiliary variable $\mathbf{T}_i$ defined as the $\mathrm{SO}(3)$ projection of $\mathbf{Y}_i^{ori}\mathbf{\Lambda}_i$, and the \textit{Bayesian flow distribution} on $\mathbf{T}_i$ is given by
    \begin{equation} \label{eq:ori_bf}
        p_F(\mathbf{T}_i \mid \mathbf{O}, \mathcal{P}; t_i) = \mathcal{M}(\mathbf{T}_i; \mathbf{O} \boldsymbol{\Lambda}^2_i).
    \end{equation}

\end{proposition}

Next, we define the receiver distribution as:
\[
p_R(\mathbf{Y}^{ori}_i \mid \boldsymbol{\theta}^{ori}_{i-1}, \mathcal{P}; t_i) = \mathbb{E}_{\hat{\mathbf{O}} \sim p_O} \left[ p_S(\mathbf{Y}^{ori}_i \mid \hat{\mathbf{O}}; \boldsymbol{\Lambda}_i) \right],
\]
\[
\text{where} \quad p_O(\hat{\mathbf{O}} \mid \boldsymbol{\theta}_{i-1}^{ori}, \mathcal{P}; t_i) = \Phi^{ori}(\mathbf{T}^{ori}_{i-1}, \mathcal{P}, t_i).
\] Here, $\Phi^{ori}$ denotes a neural network conditioned on protein target $\mathcal{P}$, parameters $\mathbf{T}_{i-1}$ and time step $t_i$. Since both Matrix Fisher distributions of $p_S$ and $p_R$ are isotropic around the rotation matrix $\mathbf{O}$, the loss function can be defined as:
\begin{align}
&L_n^{ori}(\mathbf{O}, \mathcal{P}) = n \, \mathbb{E}_{i \sim U(1, n),\, \mathbf{T}_{i-1} \sim p_F} \, D_{\mathrm{KL}}(p_S \, \| \, p_R) \nonumber \\
&= n \, \mathbb{E}_{i \sim U(1, n),\, \mathbf{T}_{i-1} \sim p_F} \, \lambda_i a(\lambda_i) \big( 3 - \mathrm{tr}(\hat{\mathbf{O}}^\top\mathbf{O})\big),
\end{align}
where $a(\lambda_i)$ is a scalar depending only on $\lambda_i$. Please refer to Sec.~\ref{app:intro_matrix_fisher} in the Appendix for more details. 

For the inference process, we generate the final residue rotations $\mathbf{O}$ by sampling from the prior $\mathbf{T}_0 \sim \mathrm{Uniform}(\mathrm{SO}(3))$ and evolving through discrete time steps $i \in \{0, 1, \dots, N\}$:
\[
\mathbf{T}_{0} 
\xrightarrow{\Psi^{ori}} \hat{\mathbf{O}} 
\xrightarrow{P_F} \mathbf{T}_1 
\rightarrow\cdots \rightarrow
\mathbf{T}_N 
\xrightarrow{\Psi^{ori}} \mathbf{O}_\text{final}.
\]

\subsection{Euclidean BFN for Centroids and Categorical BFN for Types}
In this section, we briefly describe the construction of the Bayesian flow for Gaussian centroids and discrete residue types, mainly following the methodology of previous works~\cite{graves2023bayesian, qu2024molcraft}.

The centroids (i.e., $C_\alpha$ positions) of $N$ residues in a peptide, denoted as $\mathbf{X}$, can be characterized by Gaussian distribution, i.e., $\mathbf{X}=\{\mathbf{x}^{(i)}\}_{i=1}^N \sim p_I(\cdot\mid \boldsymbol{\theta}^{pos}) = \mathcal{N}(\mathbf{X}\mid \boldsymbol{\mu}^{pos}, (\rho^{pos})^{-1}\mathbf{I})$, where $\boldsymbol{\theta}^{pos}\stackrel{\mathrm{def}}{=} \{\boldsymbol{\mu}^{pos}, \rho^{pos}\} $. 

The amino acid types in a peptide can be denoted as $C = \{c^{(i)}\}_{i=1}^{D}$ where $c^{(i)} \in \{ c \in \mathbb{Z}^+ \mid 1 \leq c \leq 20 \} $. Here, we utilize a categorical distribution $\boldsymbol{\theta}^{type}=(\boldsymbol{\theta}^{(1)},\cdots, \boldsymbol{\theta}^{(D)})$ with $\boldsymbol{\theta}^{(d)} = \big(\theta^{(d)}_1, \cdots, \theta^{(d)}_K\big) \in \Delta^{K-1}$, where $\theta^{(d)}_k$ is the probability assigned to class k for $d$-th residue type. Then the input distribution gives $p_I(C\mid\boldsymbol{\theta}^{type})=\prod_{d=1}^D \theta^{(d)}_{c^{(d)}}$. Other key components for constructing a complete Bayesian flow, including the sender and receiver distributions, Bayesian flow distributions and loss functions are summarized in Sec.~\ref{app:centroids} in the Appendix.

The final centroids $\mathbf{X}_\text{final}$ and residue types $\mathbf{C}_\text{final}$ can be sampled by initializing from simple priors. Specifically, centroids are initialized with parameters $\boldsymbol{\theta}_0^{pos} = \mathbf{0} \in \mathbb{R}^{3\times3}$, and residue types with $\boldsymbol{\theta}_0^{type} = \tfrac{1}{K}\mathbf{1}$. Both parameters follow similar Bayesian update trajectories:
\begin{align*}
\boldsymbol{\theta}_0^{pos} 
&\xrightarrow{\Psi^{pos}} \hat{\mathbf{X}} 
\xrightarrow{P_F} \boldsymbol{\theta}_1^{pos}
\rightarrow \cdots \rightarrow
\boldsymbol{\theta}_N^{pos}
\xrightarrow{\Psi^{pos}} \mathbf{X}_\text{final}, \\
\boldsymbol{\theta}_0^{type} 
&\xrightarrow{\Psi^{type}} \hat{\mathbf{C}} 
\xrightarrow{P_F} \boldsymbol{\theta}_1^{type}
\rightarrow \cdots \rightarrow
\boldsymbol{\theta}_N^{type}
\xrightarrow{\Psi^{type}} \mathbf{C}_\text{final}.
\end{align*}

\subsection{Summary of PepBFN}
By integrating all multimodal Bayesian flow objectives, the overall loss function is defined as follows:
\begin{equation} \label{eq:all_loss}
    \mathcal{L}_n^{all} = \lambda_1\mathcal{L}_n^{pos} + \lambda_2\mathcal{L}_n^{ori} + \lambda_3\mathcal{L}_n^{type} + \lambda_4\mathcal{L}_n^{ang},
\end{equation}
where $\lambda_1, \lambda_2, \lambda_3, \text{and} \ \lambda_4$ represent the respective weights of each component in the loss function. These four modality-specific modules are coupled through a single SE(3)-aware neural network to predict a denoised peptide $\mathcal{\hat{G}}$ under the Bayesian flow framework. We next evaluate this unified pipeline on side-chain packing, reverse folding, and binder co-design tasks. We provide complete training and sampling details of PepBFN in Sec.~\ref{sec:train_test} in the Appendix.

\section{Experiments}
In this section, we conduct a comprehensive evaluation of PepBFN across three tasks: (1) side-chain packing, (2) peptide reverse folding, and (3) sequence-structure co-design. Following previous works \cite{li2024pepflow, li2024hotspot}, we curated moderate-length sequences from PepBDB \cite{wen2019pepbdb} and Q-BioLip \cite{wei2024q}. Our dataset consists of 158 complexes for testing, and 8,207 samples designated for training and validation.

\subsection{Side-chain Packing} \label{sec:scpack}
This task aims to predict the side-chain torsion angles of each peptide, given fixed backbone structures and sequences. For evaluation, each model generates 64 side-chain conformations per peptide to recover the target angles. Our model PepBFN\_sc is trained under the same fixed backbone and sequence settings.

\renewcommand{\arraystretch}{1.2}
\begin{table}[h]
    \centering
\resizebox{\linewidth}{!}{
    \begin{tabular}{lccccc}
        \toprule
        \multirow{2}{*}{} & \multicolumn{4}{c}{MAE (deg)$ \downarrow$} & \multirow{2}{*}{Correct \% $\uparrow$} \\
        \cline{2-5}
         & $\chi_1$ & $\chi_2$ & $\chi_3$ & $\chi_4$ &  \\
         \hline
         Rosetta & 38.31 &43.23 &53.61 &71.67 &57.03 \\
         SCWRL4 &30.06 &40.40 &49.71 &53.79 &60.54 \\
        DLPacker &22.44 &35.65 &58.53 &61.70 & 60.91 \\
        AttnPacker &19.04 &28.49 &40.16 &60.04 &61.46 \\
        DiffPack &17.92 &26.08 &36.20 &67.82 &62.58 \\
        PepBFN\_sc & \textbf{10.75} & \textbf{12.26} & \textbf{34.25} & \textbf{53.21} & \textbf{75.24}\\ 
        \bottomrule
    \end{tabular}
}
    \caption{Evaluation of methods in side-chain packing task.}
    \label{tab:sc_packing}
    \vspace{-1em}
\end{table}

\textbf{Metrics} We evaluate performance using the mean absolute error (MAE) over the four predicted torsion angles. Given the inherent flexibility of side-chains, we additionally report the proportion of correct predictions whose deviations fall within $20^\circ$ of the ground truth.

\textbf{Baselines} Two traditional energy-based methods: Rosetta Packer~\cite{leman2020macromolecular}, SCWRL4~\cite{krivov2009improved}, and
three learning-based models: DLPacker~\cite{misiura2022dlpacker},
AttnPacker~\cite{mcpartlon2023end}, DiffPack~\cite{zhang2023diffpack}.

\textbf{Results}
As shown in Table~\ref{tab:sc_packing}, our method consistently outperforms all baselines in the side-chain packing task. 
It achieves the lowest MAE for the four torsion angles. Furthermore, our method achieves the highest proportion of correctly predicted side-chains within $20^\circ$ of the ground truth, which exceeds the second-best method by a substantial margin. 
A key insight is that our neural network directly takes $K$ Gaussian parameters as expressive inputs and yields accurate predictions of the angles. Subsequent Bayesian updates then leverage these predictions to  refine the posterior distribution while simultaneously preserving and sharpening its multi-peak structure. This integration of an expressive Gaussian mixture prior with principled Bayesian inference yields more stable and accurate torsion angle predictions than unimodal alternatives.

\subsection{Peptide Reverse Folding}
This task designs peptide sequences from backbone-only complex structures. Each model generates 64 sequences per peptide. We trained PepBFN\_seq with fixed backbones to predict the sequences.

\renewcommand{\arraystretch}{1.2}
\begin{table}[htbp]
\centering
\resizebox{\linewidth}{!}{
\begin{tabular}{lcccc}
\toprule
Method & AAR \% $\uparrow$ & Worst \% $\uparrow$ & Likeness error $\downarrow$ & Hamming Diversity $\uparrow$\\
\midrule
ProteinMPNN              & 53.28 & 45.99 & 2.70 & 15.33 \\
ESM-IF                   & 43.51 & 36.18 & 2.67 & 13.76 \\
PepFlow                  & 53.98 & 43.32 & 0.59 & \textbf{26.40} \\
PepBFN\_seq              & \textbf{62.46} & \textbf{47.82} & \textbf{0.45}  & 13.21 \\

\bottomrule
\end{tabular}
}
\vspace{-.5em}
\caption{Evaluations of methods in reverse folding task. }
\label{tab:fix_backbone}
\end{table}

\textbf{Baselines} We use three baselines for peptide sequence design: 
ProteinMPNN~\cite{dauparas2022proteinmpnn}, 
ESM-IF~\cite{hsu2022learning}, 
and PepFlow (partial sampling)~\cite{li2024pepflow}.

\textbf{Metrics} The amino acid recovery rate (\textbf{AAR}) measures the sequence identity between the generated and the ground truth. \textbf{Worst} denotes, for each protein target, the lowest AAR among the generated peptides. \textbf{Likeness error} is assessed via the mean absolute errors of negative log-likelihood computed by ProtGPT2~\cite{ferruz2022protgpt2} between peptides in test set and the generated peptides, which captures how well the generated sequences conform to the native peptide distribution. \textbf{Hamming Diversity} is evaluated as the mean pairwise Hamming distance among the generated sequences.

\textbf{Results} As shown in Table~\ref{tab:fix_backbone}, our method significantly outperforms all baselines in terms of AAR and the worst rate recovery across generated sequences, demonstrating its strong capability to recover peptide sequences under fixed backbone constraints. Moreover, our model achieves the smallest likeness error score compared to other methods, indicating that the generated sequences align most closely with native peptide distributions. The model exhibits comparatively low sequence diversity, which is expected since its high recovery rate restricts the exploration of sequence space. Overall, by modeling discrete data in continuous parameter space, our approach effectively captures the underlying data manifold,  which in turn leads to improved performance.

\subsection{Sequence-structure Co-design}

\renewcommand{\arraystretch}{1}
\begin{table*}[h]
    \centering
\resizebox{\linewidth}{!}{
    \begin{tabular}{cccccccc}
        \toprule
        \multirow{2}{*}{Experiments} & \multicolumn{3}{c}{Geometry} & \multicolumn{2}{c}{Energy} & \multicolumn{2}{c}{Design} \\
        \cmidrule(r){2-4}
        \cmidrule(r){5-6}
        \cmidrule(l){7-8}
        \multicolumn{1}{c}{} & RMSD \AA \ & SSR \% $\uparrow$ & BSR \% $\uparrow$ & Affinity \% $\uparrow$ & Stability \% $\uparrow$ & Novelty \% $\uparrow$ & Diversity \% $\uparrow$ \\
        \midrule
        RFDiffusion      &4.17 &63.86 &26.71 &16.53 &\textbf{26.82} &53.74 &25.39 \\
        ProteinGenerator &4.35 &29.15 &24.62 &13.47 &23.84 &52.39 &22.57 \\
        PepFlow          &2.07 &83.46 &86.89 &21.37 &18.15 &50.26 &20.23 \\
        PepGLAD          &3.83 &80.24 &19.34 &10.47 &20.39 &75.07 &32.10 \\
        PepHAR           &2.68 &\textbf{84.91} &86.74 &20.53 &16.62 &79.11 &29.58 \\
        PepBFN           & 3.58    &  80.40    &  \textbf{86.97}     &  \textbf{22.26}    &  17.56    &  \textbf{79.79}    &  \textbf{32.47}    \\
        \bottomrule
    \end{tabular}
}   
    \vspace{-.5em}
    \caption{Evaluation of six different methods for the peptide de novo design task.}
    \label{tab:benchmark}
    \vspace{-1em}
\end{table*}

This task involves generating both the sequence and the binding gesture of a peptide given its target protein. Each model takes the full-atom structure of the target protein as input and outputs peptide binder structures. For evaluation, we generate 64 peptides for each target protein.

\textbf{Metrics}  
 (1) \textbf{Geometry}. \textbf{RMSD (Root-Mean-Square Deviation)} computes the $C_\alpha$ distances between the generated peptide structures and the native structures after alignment. \textbf{SSR (Secondary Structure Ratio)} calculates the overlap in secondary structures between the generated and native peptides. \textbf{BSR (Binding Site Rate)} quantifies the similarity in peptide-target interactions by evaluating the overlap of the binding sites. (2) \textbf{Energy}. \textbf{Affinity} evaluates the fraction of peptides that exhibit lower binding energies compared to the native peptide. \textbf{Stability} assesses the percentage of generated complexes that exhibit lower total energy than native complexes, using the Rosetta energy function \cite{chaudhury2010pyrosetta}. (3) \textbf{Novelty} is measured as the ratio of novel peptides, defined by two criteria: 
(a) TM-score $\leq 0.5$~\cite{zhang2005tm,xu2010tm5} and 
(b) sequence identity $\leq 0.5$. 
(4) \textbf{Diversity} quantifies the structural and sequence variability among generated peptides for each target, 
computed as pairwise $(1-\text{TM-score}) \times (1-\text{sequence identity})$.

\textbf{Baselines}  
We evaluate PepBFN against five powerful peptide design models. \textbf{RFDiffusion} \cite{watson2023novo} leverages pre-trained weights from RoseTTAFold \cite{baek2021accurate} to generate protein backbone structures through a denoising diffusion process. The peptide sequences are subsequently reconstructed using \textbf{ProteinMPNN} \cite{dauparas2022proteinmpnn}. \textbf{ProteinGenerator} enhances RFDiffusion by incorporating joint sequence-structure generation \cite{lisanza2023pg}. \textbf{PepFlow} \cite{li2024pepflow} generates full-atom peptides using a flow matching framework. \textbf{PepGLAD} \cite{kong2024pepglad} utilizes equivariant latent diffusion networks to generate full-atom peptide structures. \textbf{PepHAR} \cite{li2024hotspot} generates peptide residues autoregressively, based on a learned prior distribution for hotspot residues.

 \begin{figure}[h]
    \centering
    \includegraphics[width=0.8\linewidth]{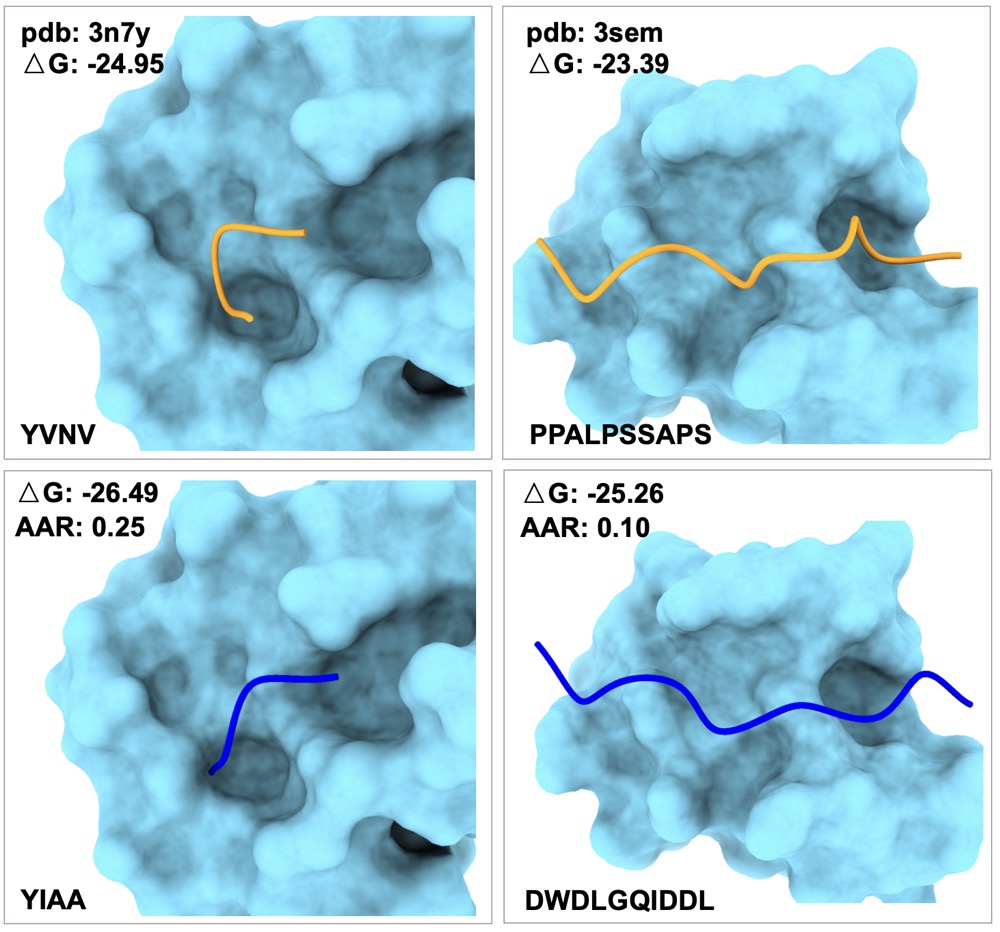}
    \vspace{-.5em}
    \caption{Two examples of PepBFN-generated peptides with improved binding affinities. Top row: native peptides; bottom row: peptides generated by our method.
}
    \label{fig:vis}
\end{figure}

\textbf{Results} As shown in Table~\ref{tab:benchmark}, PepBFN effectively generates diverse and novel peptides with strong binding affinities. Specifically, PepBFN achieves the best binding affinity (22.26\%) and binding site rate (86.97\%), indicating effective modeling of peptide–protein interactions. In addition, it also achieves the highest novelty (79.79\%) and diversity (32.47\%). These improvements can be attributed to the smooth and expressive generative process of BFN, which accurately models the underlying parameter manifold and facilitates diverse yet plausible peptide generation. Interestingly, RFDiffusion exhibits better stability, probably because it was trained on a large corpus of proteins with more stable motifs~\cite{li2024pepflow}. Compared with PepFlow and PepHAR, our method achieves lower binding energies despite exhibiting slightly higher RMSD values. This suggests that achieving strong binding affinity does not require minimal RMSD, as moderate deviations, particularly outside the binding interface, do not significantly disrupt critical binding interactions. As shown in Fig.~\ref{fig:vis}, our method can generate peptides with different binding gestures while maintaining better binding affinities.

Together, experiments across three distinct tasks highlight the advantages of fully continuous parameterization for multimodal peptide modeling and underscore the effectiveness of the PepBFN framework.

\begin{figure}[htbp]
  \centering
  \begin{subfigure}[t]{0.48\linewidth}
    \includegraphics[width=\linewidth]{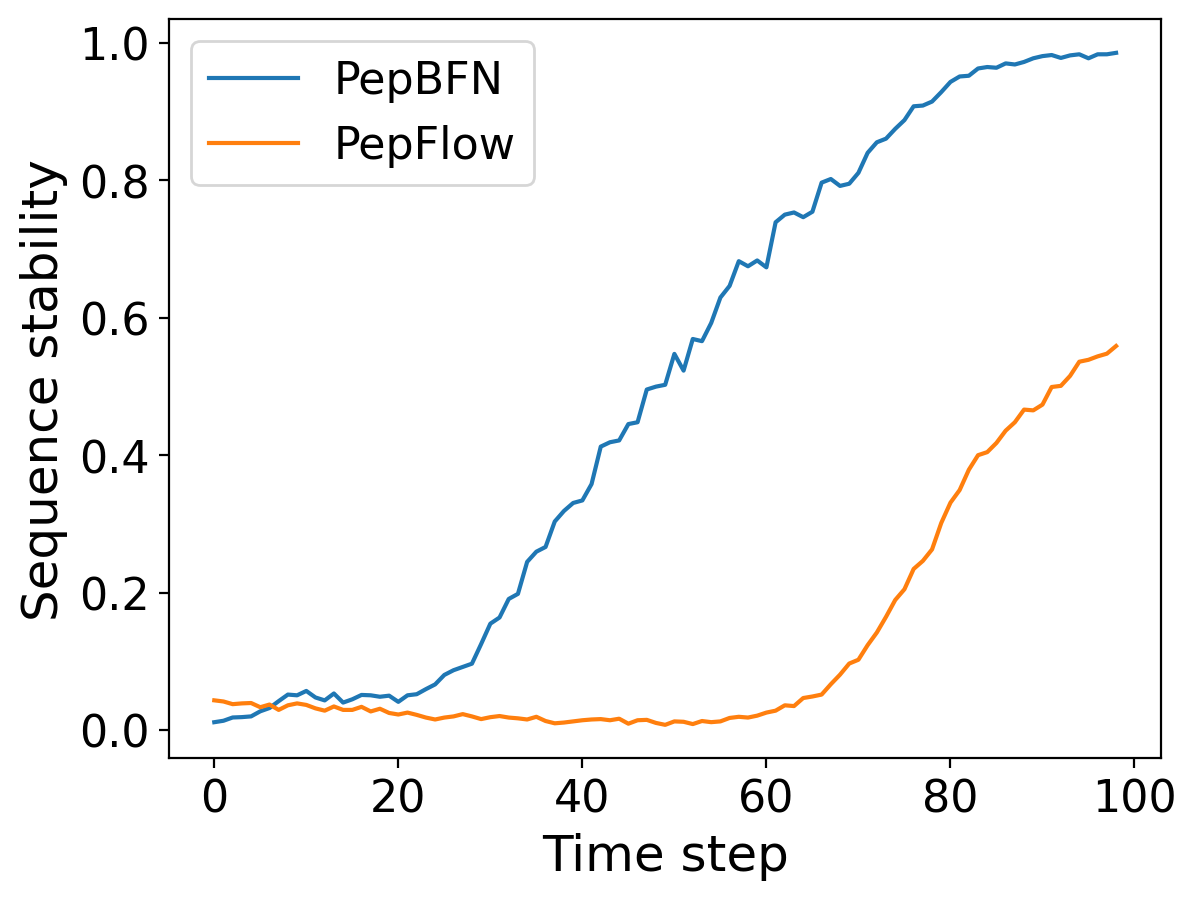}
    \vspace{-1.5em}
    \caption{}
    \vspace{-1em}
    \label{fig:BFN_VS_FLOW}
  \end{subfigure}
  \hfill
  \begin{subfigure}[t]{0.48\linewidth}
    \includegraphics[width=\linewidth]{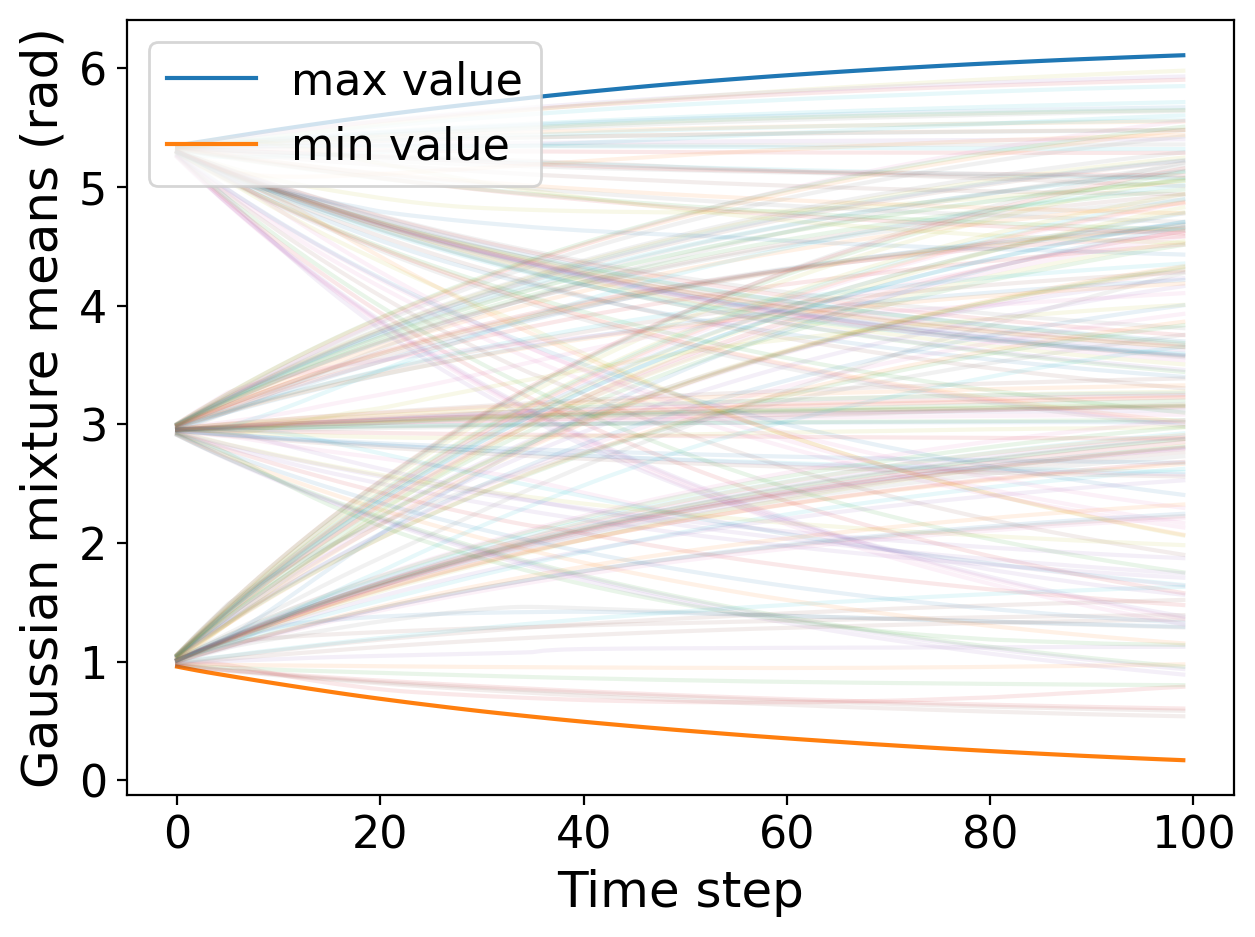}
    \caption{}
    \vspace{-1em}
    \label{fig:GMM_component_means}
  \end{subfigure}
  \caption{(a) Sequence stability of peptides during generation in peptide binder design task, measured by the fraction of sequences that no longer change. (b) Trajectories of Gaussian mixture component means for side-chain torsion angles.}
  \label{fig:main_comparison}
  \vspace{-1em}
\end{figure}

\section{Ablation Studies}

\subsection{Fast Convergence and High Sequence Stability}  
We first investigate whether modeling continuous-discrete data distributions in fully continuous parameter space can alleviate the aforementioned data-type mismatch. As shown in Fig.~\ref{fig:BFN_VS_FLOW}, PepBFN converges significantly faster and achieves higher sequence stability than PepFlow, which directly models peptide distributions in hybrid continuous-discrete data space. This finding confirms that a fully continuous parameterization helps resolve the mismatch issue and facilitates faster, more stable generation.

\renewcommand{\arraystretch}{1.2}
\begin{table}[h]
    \centering
\resizebox{\linewidth}{!}{
    \begin{tabular}{lccccc}
        \toprule
        \multirow{2}{*}{} & \multicolumn{4}{c}{MAE (deg) $\downarrow$} & \multirow{2}{*}{Correct \% $\uparrow$} \\
        \cline{2-5}
         & $\chi_1$ & $\chi_2$ & $\chi_3$ & $\chi_4$ &  \\
         \hline
        PepBFN\_uni &24.50 & 32.44 & 56.22 & 60.35 & 59.54 \\
        PepBFN\_sc & \textbf{10.75} & \textbf{12.26} & \textbf{34.25} & \textbf{53.21} & \textbf{75.24}\\ 
        \bottomrule
    \end{tabular}
}   
    \caption{Unimodal Gaussian vs Gaussian mixture in side-chain packing task.}
    \label{tab:abla_sc_packing}
\end{table}

\subsection{Effective Gaussian Mixture-based Modeling} To further evaluate the role of Gaussian mixture distributions in modeling side-chain torsions, we replace the mixture with a single Gaussian (PepBFN\_uni) while keeping all other components unchanged. Both PepBFN\_uni and PepBFN\_sc are trained with fixed backbones and sequences (without side-chain contexts) under the Bayesian flow framework. As shown in Table~\ref{tab:abla_sc_packing}, replacing the Gaussian mixture distributions with a single Gaussian significantly reduces side-chain prediction accuracy. In addition, as shown in Fig.~\ref{fig:GMM_component_means}, although no explicit periodicity constraint was applied, the initialization, sampling, and Bayesian update mechanism naturally constrain all Gaussian mixture component means within the $[0, 2\pi]$ interval, implicitly preserving the periodic nature of torsion angles.

\section{Conclusion}

We introduce PepBFN, the first Bayesian Flow Network designed for full-atom peptide generation. Our Gaussian mixture-based Bayesian flow framework effectively captures the multimodal nature of side-chain torsion angle distributions. Besides, by jointly modeling residue orientations, torsional angles, centroids, and amino acid types in fully continuous parameter space, PepBFN enables smooth parameter updates and exhibits stable peptide sequence generation. Evaluated across three peptide design benchmarks, PepBFN consistently outperforms existing methods. While this work establishes the potential of Bayesian flow networks for peptide design, there remains room for improvement. For instance, incorporating backbone generation in earlier steps and modeling sequence and side-chain generations in later steps may reduce the complexity of the task and further enhance performance. Overall, PepBFN establishes a principled and unified Bayesian flow framework for peptide design, providing a foundational tool to advance peptide engineering.

\section*{Acknowledgments}
This work was supported by the National Natural Science Foundation of China (grants No. 62172273), the Science and Technology Commission of Shanghai Municipality (grants No. 24510714300), and the Shanghai Municipal Science and Technology Major Project, China (Grant No. 2021SHZDZX0102).

\bibliography{aaai2026}

\clearpage

\setcounter{section}{0}
\twocolumn[
  \begin{center}
    {\Huge \bfseries Appendix} 
    \vspace{1em}
  \end{center}
]

\renewcommand{\arraystretch}{2.5}
\begin{table*}[h]
\begin{center}
\resizebox{\linewidth}{!}{
\begin{tabular}{ccccc}
\toprule
Method &  State dynamics & Generation process &Neural network output $\phi(\cdot)$ & Loss function at $t$ time step \\
\hline
DDPM & $q(x_t\mid x_1)$ &   $p_{\theta}(x_{t+\triangle t}\mid \hat{x}_1, x_t)$ & $\hat{x}_1 = \phi(x_t)$ & $\mathrm{KL}\big(q(x_{t+\triangle t}\mid x_1, x_t) \ \| \ p_{\theta}(x_{t+\triangle t}\mid \hat{x}_1, x_t\big)$  \\
\hline
Flow matching & $x_t = (1-t)x_0 + tx_1$& $\mathrm{ODE}(\hat{x}_1-x_0,t)$ & $\hat{v} = \hat{x}_1 - x_0 = \phi(x_t)$ & $\|(x_1-x_0, \hat{x}_1-x_0)\|_\mathrm{x-norm}$\\
\hline
BFN & $p_F(\theta_t \mid x; t)= p_U(\theta_t \mid \theta_0, x, t)$
& $\begin{aligned}
    \theta_{t+\triangle t} & = \mathrm{h}(\theta_t, y, \alpha_t), \\
     \text{where}~y&\sim  p_R(\cdot\mid \hat{x}, \alpha_t)
\end{aligned}$
& $\hat{x} = \phi(\theta_t)$ & $\mathrm{KL}\big(p_S(y\mid x, \alpha_t) \ \| \ p_R(y \mid \hat{x}, \alpha_t)\big)$\\
\bottomrule
\end{tabular}
}
\caption{Comparison of Generative Modeling Frameworks: DDPM, Flow Matching, and BFN. 
For clarity, we denote $x_1$ as the data sample, $x_0$ as noise drawn from the prior distribution and $\theta_0$ as the initial parameter. Here, $\mathrm{x\text{-}norm}$ refers to the norm defined on the manifold where $x$ resides.
}
\label{tab_app:compare}
\end{center}
\end{table*}

\section{An Introduction on Matrix Fisher Distribution}\label{app:intro_matrix_fisher}

\subsection{Probability Density
Function of Matrix Fisher Distribution}
The Matrix Fisher distribution is a probability distribution over the rotation group $\mathrm{SO}(3)$, which represents 3D rotation matrices. A random rotation matrix $R \in \mathrm{SO}(3)$ is said to follow a Matrix Fisher distribution if its probability density function (PDF) is given by:
\[
\mathcal{M}(R; F) = \frac{1}{c(F)} \exp\left( \mathrm{tr}(F^\top R) \right),
\]
where $F \in \mathbb{R}^{3 \times 3}$ is the matrix parameter, $\mathrm{tr}(\cdot)$ denotes the trace operator, and $c(F)$ is a normalizing constant ensuring that the density integrates to one over $\mathrm{SO}(3)$. The parameter matrix $F$ can be decomposed as $F = U \Lambda V^\top$, where $U, V \in \mathrm{SO}(3)$ are rotation matrices and $\Lambda = \mathrm{diag}(\lambda_1, \lambda_2, \lambda_3)$ is a diagonal matrix of concentration parameters. The mode of the distribution is given by $R = UV^\top$, and the concentration around the mode is controlled by the magnitudes of $\lambda_i$.

\subsection{Derivation of the KL Divergence Between Two Isotropic Matrix Fisher Distributions} \label{app:sec_mfkl}

Consider two Matrix Fisher distributions on $\mathrm{SO}(3)$ with identical diagonal concentration parameters $\Lambda = \lambda I_3$, but different mean orientations $X, X' \in \mathrm{SO}(3)$:
\[
p_A(R) = \mathcal{M}(X\lambda I_3), \qquad
p_B(R) = \mathcal{M}(X'\lambda I_3).
\]
Their density functions are given by:
\[
p(R;F) = \frac{1}{c(F)} \exp\big( \mathrm{tr}(F^\top R) \big), \qquad F = X\lambda I_3.
\]
The KL divergence between $p_A$ and $p_B$ is:
\[
D_{KL}(p_A||p_B) = \mathbb{E}_{R \sim p_A}[\log p_A(R) - \log p_B(R)].
\]
Since the normalization constant $c(F)$ depends only on the singular values of $F$ (which are identical for $p_A$ and $p_B$), the ratio of partition functions cancels, yielding:
\[
D_{KL}(p_A||p_B) = \lambda \, \mathbb{E}_{R \sim p_A} \big[ \mathrm{tr}(X^\top R) - \mathrm{tr}(X'^\top R) \big].
\]

Let $Q = X^\top R$. By the rotational equivariance property of the Matrix Fisher distribution, if $R \sim \mathcal{M}(X\lambda I_3)$, then:
\[
Q \sim \mathcal{M}(\lambda I_3).
\]
Since $R = XQ$, we have:
\[
\mathrm{tr}(X^\top R) = \mathrm{tr}(Q),
\qquad
\mathrm{tr}(X'^\top R) = \mathrm{tr}(\Delta^\top Q),
\]
where $\Delta = X^\top X'$ is the relative rotation between the two mean orientations.  
Thus:
\[
D_{KL}(p_A||p_B) = \lambda \, \mathbb{E}_{Q \sim \mathcal{M}(\lambda I_3)} \big[ \mathrm{tr}(Q) - \mathrm{tr}(\Delta^\top Q) \big].
\]

By linearity of the trace and expectation:
\[
D_{KL}(p_A||p_B) = \lambda \Big( \mathrm{tr}(\mathbb{E}[Q]) - \mathrm{tr}(\Delta^\top \mathbb{E}[Q]) \Big),
\]
where the expectation is over $Q \sim \mathcal{M}(\lambda I_3)$.  
For the isotropic case, the first moment is proportional to the identity:
\begin{equation} \label{eq:a}
\mathbb{E}[Q] = a(\lambda) I_3,
\end{equation}
where $a(\lambda)$ is a scalar depending only on $\lambda$. This coefficient can be obtained numerically via sampling:
\[
a(\lambda) = \frac{1}{3}\mathrm{tr}(\mathbb{E}[Q]).
\]

\begin{figure}
    \centering
    \includegraphics[width=0.8\linewidth]{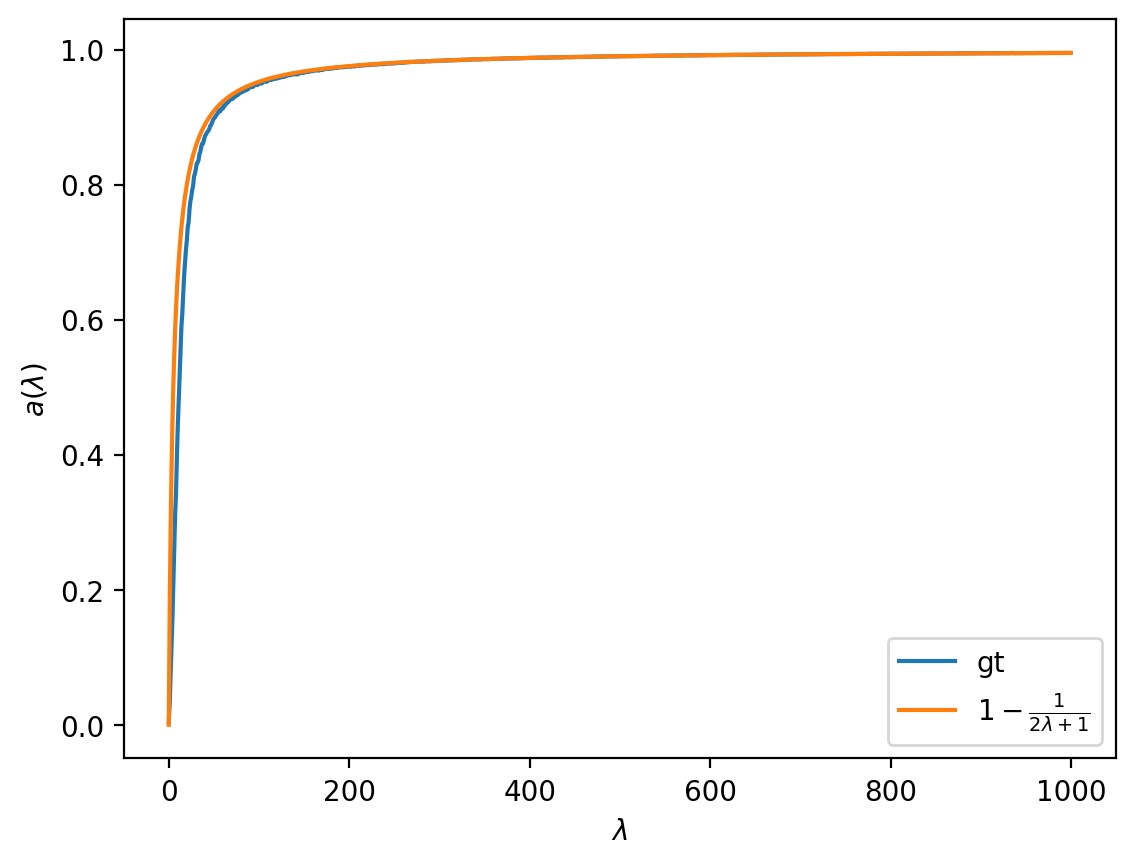}
    \caption{$a(\lambda)$ vs. $\lambda$}
    \label{fig:avslambda}
\end{figure}

Substituting $\mathbb{E}[Q] = a(\lambda) I_3$, we obtain:
\[
D_{KL}(p_A||p_B) = \lambda a(\lambda) \big( \mathrm{tr}(I_3) - \mathrm{tr}(\Delta) \big).
\]
Since $\mathrm{tr}(I_3)=3$, the closed-form KL divergence is:
\[
D_{KL}(p_A||p_B) = \lambda a(\lambda) \big( 3 - \mathrm{tr}(\Delta) \big),
\]
where $\Delta = X^\top X'$ represents the relative rotation between the two mean orientations. 
As illustrated in Fig.~\ref{fig:avslambda}, the approximation 
\begin{equation} \label{eq:lambda}
    a(\lambda) \approx 1 - \frac{1}{2\lambda+1}
\end{equation}
fits the empirical values well, and we adopt this approximation in practice.

Note that $\mathrm{tr}(\Delta)$ can be expressed in terms of the relative rotation angle $\theta$ as $\mathrm{tr}(\Delta) = 1 + 2\cos\theta$, leading to:
\[
D_{KL}(p_A||p_B) = 2\lambda a(\lambda)(1-\cos\theta).
\]
This formulation shows that the KL divergence depends between $p_A$ and $p_B$ only on the relative angle $\theta$ between $X$ and $X'$ and the scalar concentration parameter $\lambda$.

\subsection{Fast Sampling for Matrix Fisher Distribution} \label{app:fast_sample_matrix}

\begin{figure*}[htbp]
    \centering
    \begin{subfigure}[b]{0.95\linewidth}
        \centering
        \includegraphics[width=\linewidth]{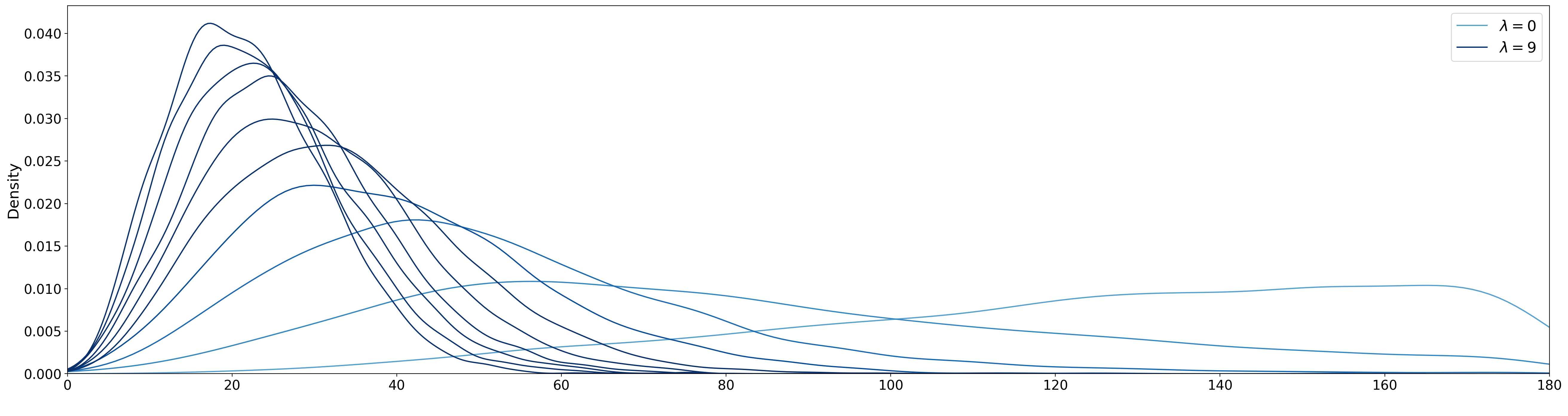}
        \vspace{-2em}
        \caption{}
        \label{fig:lambda0_9}
    \end{subfigure}
    \begin{subfigure}[b]{0.95\linewidth}
        \centering
        \includegraphics[width=\linewidth]{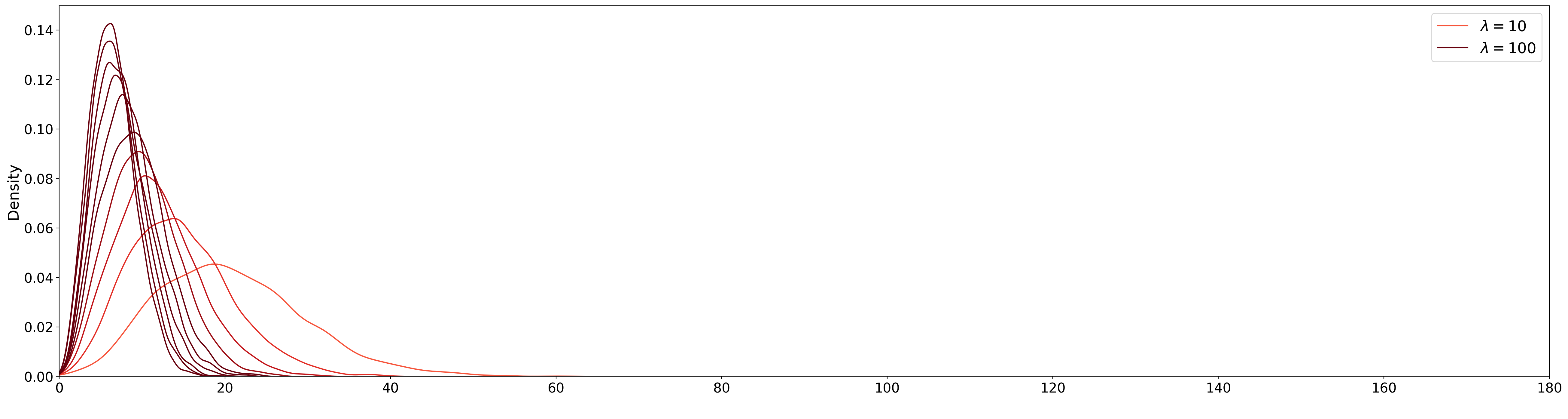}
        \vspace{-2em}
        \caption{}
        \label{fig:lambda10_100}
    \end{subfigure}
    \caption{Distributions of rotation angles implied by Matrix Fisher distributions for varying concentration parameters \(\lambda\). Panel (a) shows the low \(\lambda\) regime (\(\lambda=0\) to 9) in blue, where the angle distributions are broad and shift as \(\lambda\) increases. Panel (b) depicts higher \(\lambda\) values (10 to 100) in red, illustrating progressive sharpening and localization of the rotation angle mass as the concentration grows.}
    \label{fig:matrix_fisher_angles}
\end{figure*}

To sample from the isotropic Matrix Fisher distribution $\mathcal{M}(\lambda I)$ on $\mathrm{SO}(3)$, we adopt a hybrid strategy based on the concentration parameter $\lambda$. For small $\lambda\leq 26$, we use rejection sampling from the uniform distribution over $\mathrm{SO}(3)$, accepting samples based on the unnormalized Matrix Fisher density. For large $\lambda$, we approximate the distribution using a Gaussian over the Lie algebra $\mathfrak{so}(3)$, followed by exponentiation via the Rodrigues' formula to map to $\mathrm{SO}(3)$ (see Fig.~\ref{fig:matrix_fisher_angles}). 

\begin{algorithm}[H]
\caption{Hybrid Sampling from Matrix Fisher $\mathcal{M}(\lambda I)$}
\label{alg:mf_sampling}
\begin{algorithmic}[1]
\State \textbf{Input:} parameter $\lambda$, number of samples $N$
\State \textbf{Output:} sampled rotation matrices $\{R_i\}_{i=1}^N \in \mathrm{SO}(3)$
\If{$\lambda \leq 26$}
    \State Set maximum density $d_{\max} \gets \exp(3\lambda)$
    \State Initialize sample set $\mathcal{R} \gets \emptyset$
    \While{$|\mathcal{R}| < N$}
        \State Sample $B$ rotations $R \sim \text{Uniform}(\mathrm{SO}(3))$
        \State Compute density: $d_i = \exp(\lambda \cdot \text{tr}(R_i))$
        \State Accept $R_i$ with probability $d_i / d_{\max}$
        \State Add accepted $R_i$ into $\mathcal{R}$
    \EndWhile
    \State $R_{1:N} \gets$ first $N$ samples in $\mathcal{R}$
\Else
    \State Compute $\sigma \gets \frac{1}{\sqrt{2\lambda}}$
    \State Sample $\omega_i \sim \mathcal{N}(0, \sigma^2 I_3)$ for $i = 1, \dots, N$
    \State Map to $\mathrm{SO}(3)$: $R_i \gets \exp_{\mathrm{SO}(3)}(\omega_i)$
\EndIf
\State \textbf{Return:} $\{R_i\}_{i=1}^N$
\end{algorithmic}
\end{algorithm}

\section{Propositions and Proofs}

\subsection{Proof of Lemma~\ref{lemma:gmm}}

\begin{proof}
Consider a prior over $x$ given by a $K$-component Gaussian mixture,
\[
p(x) = \sum_{k=1}^K \pi^k \mathcal{N}(x \mid \mu^k, (\rho^k)^{-1}),
\]
where $\pi^k \ge 0$ and $\sum_{k=1}^K \pi^k = 1$, $\mu^k$ is the mean, and $\rho^k$ the precision of the $k$-th component. The likelihood is a Gaussian,
\begin{equation}
p(y \mid x) = \mathcal{N}(y \mid x, \alpha^{-1}),
\end{equation}
with precision $\alpha$.
By Bayes’ rule, the posterior is
\begin{align}
    &p(x \mid y) \propto p(y \mid x) p(x) \nonumber \\
    &= \sum_{k=1}^K \pi^k \, \mathcal{N}(y \mid x, \alpha^{-1}) \mathcal{N}(x \mid \mu^k, (\rho^k)^{-1}).
\end{align}
Applying the standard identity for the product of two Gaussians, we can rewrite the $k$-th component
\begin{align}
    &\mathcal{N}(y \mid x, \alpha^{-1}) \mathcal{N}(x \mid \mu^k, (\rho^k)^{-1}) \nonumber \\
    &= \mathcal{N}(y \mid \mu^k, \alpha^{-1} + (\rho^k)^{-1}) \,
       \mathcal{N}(x \mid \tilde{\mu}^k, (\tilde{\rho}^k)^{-1}),
\end{align}
where the first term corresponds to the marginal likelihood (normalization factor), and the second term is the posterior over $x$ for the $k$-th component, with
\begin{align}
    \tilde{\rho}^k &= \rho^k + \alpha, \\
\tilde{\mu}^k &= \frac{\rho^k \mu^k + \alpha y}{\tilde{\rho}^k} \label{eq:poster_mu}.
\end{align}
Normalizing over all $k$ yields the posterior weights,
\begin{equation}\label{eq:poster_pi}
    \tilde{\pi}^k = \frac{\pi^k \, \mathcal{N}(y \mid \mu^k, \alpha^{-1} + (\rho^k)^{-1})}{\sum_{j=1}^K \pi^j \, \mathcal{N}(y \mid \mu^j, \alpha^{-1} + (\rho^j)^{-1})}.
\end{equation}
Thus, the posterior retains the GMM form:
\begin{equation}
    p(x \mid y) = \sum_{k=1}^K \tilde{\pi}^k \mathcal{N}(x \mid \tilde{\mu}^k, (\tilde{\rho}^k)^{-1}).
\end{equation}
\end{proof}
\subsection{Proof of Proposition~\ref{prop:bfn4gmm}}
\begin{proof}
    
Consider the mean parameter of $k$-th Gaussian in a temporal context (with time step $t_i$), we can rewrite Eq.~\ref{eq:poster_mu} as:
\begin{equation}
    \mu^k_i = \frac{\rho_{i-1}^k \mu_{i-1}^k + \alpha y}{\rho^k_i} \quad \text{with} \quad \rho_i^k = \rho^k_{i-1} + \alpha.
\end{equation}
This form exactly matches the mean update in \cite{graves2023bayesian} (Eq.~50), demonstrating that each GMM component's mean evolves identically to a single Gaussian case. Applying Eq.~53 in~\cite{graves2023bayesian} for the $k$-th component of GMM, we have

\begin{equation}
    p_U^k(\mu_i^k \mid \mu_{i-1}^k, \chi; \alpha) 
= \mathcal{N}\!\left(
\mu_i^k \,\middle|\, 
\frac{\alpha \chi + \mu_{i-1}^k\rho_{i-1}^k}{\rho_i^k}, 
\frac{\alpha}{(\rho_i^k)^2}
\right)\label{eq:k_update_distribution},
\end{equation}
and the precisions are additive, i.e., $\beta(t_i) \stackrel{\text{def}}{=} \sum_{m=1}^{i} \alpha(t_m)$ if we follow the derivations in Sec. 4.4 in~\cite{graves2023bayesian}.

Recall from Eq.~\ref{eq:bayesian_flow} that 
\begin{equation}
    p_F(\boldsymbol{\theta}_i \mid \chi, \mathcal{P}; t_i) = p_U(\boldsymbol{\theta}_i \mid \boldsymbol{\theta}_0, \chi, \mathcal{P}; \beta(t_i)).
\end{equation}

Therefore, setting $\{\mu_{i-1}^k, \rho_{i-1}^k\} = \{\mu_0^k,\rho_0^k\}$ and $\alpha=\beta(t_i)$ in Eq.~\ref{eq:k_update_distribution}, and recalling that $\rho_i^k=\rho_0^k+\beta(t_i)$, we have the Bayesian flow distribution for $k$-th component:
\begin{equation}
    p_F(\mu^k_i\mid\chi,\mathcal{P};t_i) = \mathcal{N}\!\left(
\mu_i^k \,\middle|\, 
\frac{\beta(t_i) \chi + \mu_0^k\rho_0^k}{\rho_0^k+\beta(t_i)}, 
\frac{\beta(t_i)}{(\rho_0^k+\beta(t_i))^2}\right),
\end{equation}
where $\beta(t_i)$ is predefined such that the expected entropy of the input distribution $p_I(\chi \mid \boldsymbol{\theta}^{ang})$ decreases linearly over the time steps $t_i$.

The posterior weights of a Gaussian mixture distribution in Eq.~\ref{eq:poster_pi} lack closed-form solutions because the marginal likelihood integrals are analytically intractable. And thus we cannot derivate the corresponding $p_U^k(\pi^k_i\mid\pi_{i-1}^k,\chi;\alpha)$ and $p_F^k(\pi^k_i\mid\pi_0^k,\chi;\alpha)$. We propose a simulation-based procedure (Algorithm~\ref{alg:simu_GMM}) to address this problem.

\end{proof}

\subsection{Proof of Proposition~\ref{prop:upper_bound}}

\begin{proof}
    For a $K$-component Gaussian mixture distribution with identical isotropic variance $\sigma^2$ for each component,
\[
p(x) = \sum_{k=1}^K \pi_k \, \mathcal{N}(x \mid \mu_k, \sigma^2), 
\qquad \sum_{k=1}^K \pi_k = 1,
\]
the differential entropy is given by
\[
H(p) = -\int p(x) \log p(x) \, dx,
\]
which has a upper bound~\cite{huber2008entropy} :
\[
H(p) \leq H_u(x) = \sum_{k=1}^K \pi^k \cdot 
\Bigg( -\log \pi^k
+ \frac{1}{2} \log \big( 2\pi e \sigma^2 \big) \Bigg).
\]
Note that given a known time step $t_i$, the variance $\sigma^2$ is deterministic, and we have 
\begin{align}
H(t_i) & \leq \mathbb{E}_{p_F(\boldsymbol{\theta}^{ang}_i \mid \chi,\mathcal{P}; t_i)} \nonumber \\
& \Bigg[ \sum_{k=1}^K \pi^k \cdot 
\Bigg( -\log \pi^k + \frac{1}{2} \log \big( 2\pi e \sigma^2 \big) \Bigg) \Bigg]  \nonumber \\
& \leq \mathbb{E}_{p_F(\boldsymbol{\theta}^{ang}_i \mid \chi,\mathcal{P}; t_i)}  \nonumber \\
& \Bigg[  
\Bigg( \log K + \sum_{k=1}^K \pi^k \cdot \frac{1}{2} \log \big( 2\pi e \sigma^2 \big) \Bigg) \Bigg]  \nonumber \\
& = \log K + \frac{1}{2}\log(2\pi e\sigma^2) \cdot \sum_{k=1}^K \mathbb{E}_{p_F(\pi^k \mid \chi,\mathcal{P}; t_i)} \pi^k  \nonumber \\
& \leq \log K + \frac{K}{2}\log(2\pi e\sigma^2) \cdot 
\end{align}
Let $H_u(t) \stackrel{\mathrm{def}}{=} \frac{K}{2}\log(2\pi e\sigma^2)$. 
Recall that our objective is to make $H(t_i)$ decrease linearly with time step $t$. 
Since the exact entropy $H(t)$ is intractable for a Gaussian mixture-based BFN, we instead enforce its upper bound $H_u(t)$ to decay linearly with $t$ as a practical surrogate. Then we have:
\begin{align}
    H_u(t) &= (1-t)H_u(0) + tH_u(1) \\
    \Rightarrow \frac{K}{2}\log(\frac{2\pi e}{\rho_t}) &= (1-t)\cdot\frac{K}{2}\log(\frac{2\pi e}{\rho_0}) + t\cdot\frac{K}{2}\log(\frac{2\pi e}{\rho_1}) \nonumber\\
    \Rightarrow \rho_t &= \rho_0^{1-t}\cdot\rho_1^t  \nonumber \\
    \Rightarrow  \rho_0 + \beta(t) &= \rho_0^{1-t} \cdot \rho_1^t \nonumber \\
    \Rightarrow \beta(t) &= \rho_0^{1-t} \cdot \rho_1^t - \rho_0. 
\end{align}
Under discrete time steps, we have:
\begin{align}
    \alpha_i &= \beta(t_i)-\beta(t_{i-1}) \nonumber \\
    &= \rho_0^{1-i/n} \cdot \rho_1^{i/n} - \rho_0^{1-(i-1)/n} \cdot \rho_1^{(i-1)/n} \nonumber \\
    &=\rho_0^{1-i/n} \cdot \rho_1^{i/n}\Bigg[1-(\frac{\rho_0}{\rho_1})^{1/n}   \Bigg]
\end{align}
as required.
\end{proof}

\subsection{Proof of Lemma~\ref{proposition:fisher_conjugacy}}
\begin{proof}
    According to Bayesian rules, the posterior distribution
    \begin{align}
        p(\mathbf{O}\mid\mathbf{Y}_i^{ori}) &\propto p(\mathbf{O})\cdot p(\mathbf{Y}_i^{ori}\mid \mathbf{O}) \nonumber \\
        &=\exp\left(\mathrm{tr}(\boldsymbol{\theta}_a^\top \mathbf{O})\right) \cdot \exp\left(\mathrm{tr}\left((\mathbf{O\Lambda})^\top \mathbf{Y}_i^{ori}\right)\right) \nonumber \\
        &= \exp\left(\mathrm{tr}(\boldsymbol{\theta}_a^\top \mathbf{O})\right) \cdot \exp\left(\mathrm{tr}\left(\mathbf{\Lambda} \mathbf{O}^\top \mathbf{Y}_i^{ori}\right)\right) \nonumber \\
        &= \exp\left(\mathrm{tr}(\boldsymbol{\theta}_a^\top \mathbf{O})\right) \cdot \exp\left(\mathrm{tr}\left((\mathbf{Y}_i^{ori})^\top  \mathbf{O} \mathbf{\Lambda}^\top\right)\right) \nonumber \\
        &= \exp\left(\mathrm{tr}(\boldsymbol{\theta}_a^\top \mathbf{O})\right) \cdot \exp\left(\mathrm{tr}\left(\mathbf{\Lambda}^\top(\mathbf{Y}_i^{ori})^\top  \mathbf{O}  \right)\right) \nonumber \\
        &= \exp\left(\mathrm{tr}((\boldsymbol{\theta}_a+\mathbf{Y}_i^{ori}\mathbf{\Lambda})^\top \mathbf{O})\right) \nonumber \\
        &= \exp\left(\mathrm{tr}(\boldsymbol{\theta}_b^\top\mathbf{O})\right)
    \end{align}
    
\end{proof}

\subsection{Proof of Proposition~\ref{thm:fisher}}
\begin{proof}
Since both $p_I(\mathbf{O}\mid \boldsymbol{\theta}^{ori})$ and $p_S(\mathbf{Y}_i^{ori} | \mathbf{O}; \boldsymbol{\Lambda}_i)$ are Matrix Fisher distributions, Eq.~\ref{eq:bayesian_update_func} can be applied to obtain the following Bayesian update function for parameters $\boldsymbol{\theta}^{ori}_{i-1}$ and sender sample $\mathbf{Y}_i^{ori}$ drawn from $p_S(\cdot \mid \mathbf{O}; \boldsymbol{\Lambda}_i) = \mathcal{M}(\cdot; \mathbf{O} \boldsymbol{\Lambda}_i)$:

\begin{equation}
    h(\{\boldsymbol{\theta}_{i-1}^{ori}\}, \mathbf{Y}_i^{ori}, \boldsymbol{\Lambda}_{i-1}) = \{\boldsymbol{\theta}_{i}^{ori}\},
\end{equation}
with
\begin{equation}
\label{eq:update}
    \boldsymbol{\theta}_{i}^{ori} = \boldsymbol{\theta}_{i-1}^{ori} + \mathbf{Y}_i^{ori}\mathbf{\Lambda}_{i}.
\end{equation}

Eq.~\ref{eq:update} updates $\boldsymbol{\theta}_i$ based on a single sample $\mathbf{Y}_i^{ori}$ drawn from the sender distribution. When marginalizing over $\mathbf{Y}_i^{ori} \sim \mathcal{M}(\cdot; \mathbf{O}\boldsymbol{\Lambda}_i)$ as specified in Eq.~\ref{eq:bayesian_update_distribution}, the resulting distribution of $\boldsymbol{\theta}_i$ does not admit a closed-form solution, as the Matrix Fisher family is not closed under additive operations. To keep the flow on 
$\mathrm{SO}(3)$ and remain computationally tractable, we adopt a projection–refit surrogate by defining an auxiliary variable $\mathbf{T}_i$:
\begin{equation}
    \tilde{\mathbf{Y}}_i \stackrel{\mathrm{def}}{=} \mathbf{Y}_i^{ori}\mathbf{\Lambda}_i, 
    \qquad 
    \mathbf{T}_i \stackrel{\mathrm{def}}{=} \mathrm{{Proj}_{SO(3)}}(\tilde{\mathbf{Y}}_i),
    \label{eq:projection_def}
\end{equation}
where $\mathrm{{Proj}_{SO(3)}}(\cdot)$ denotes the standard projection onto the rotation manifold 
under the Frobenius norm,
\[
\mathrm{{Proj}_{SO(3)}}(\mathbf{Z}) 
= \arg\min_{\mathbf{R}\in \mathrm{SO}(3)} \|\mathbf{Z}-\mathbf{R}\|_F
= \mathbf{Z}(\mathbf{Z}^\top\mathbf{Z})^{-\frac{1}{2}}.
\]

This closed form arises from solving the orthogonal Procrustes problem,
\[
\min_{\mathbf{R}\in \mathrm{SO}(3)} \|\mathbf{Z}-\mathbf{R}\|_F^2,
\]
which seeks the rotation closest to $\mathbf{Z}$ in the Frobenius norm. 
Let $\mathbf{Z}=\mathbf{U}\mathbf{S}\mathbf{V}^\top$ be the singular value decomposition (SVD) of $\mathbf{Z}$.
The optimal rotation is then $\mathbf{R}^*=\mathbf{U}\mathbf{V}^\top$, 
and it can be equivalently written as
\[
\mathbf{R}^*=\mathbf{Z}(\mathbf{Z}^\top\mathbf{Z})^{-\frac{1}{2}},
\]
because $\mathbf{Z}^\top\mathbf{Z}=\mathbf{V}\mathbf{S}^2\mathbf{V}^\top$ 
and $(\mathbf{Z}^\top\mathbf{Z})^{-1/2}=\mathbf{V}\mathbf{S}^{-1}\mathbf{V}^\top$, 
which yields $\mathbf{U}\mathbf{V}^\top$ after substitution.
This expression ensures $\mathbf{R}^*\in \mathrm{SO}(3)$ and $\mathbf{R}^{*\top}\mathbf{R}^*=\mathbf{I}$, 
hence it is the unique closest rotation to $\mathbf{Z}$ in Frobenius distance.

When $\mathbf{\Lambda}_i = k\mathbf{I}$ for a scalar concentration $k>0$, 
we have $\tilde{\mathbf{Y}}_i = k \mathbf{Y}_i^{ori}$ with $\mathbf{Y}_i^{ori} \in \mathrm{SO}(3)$, leading to
\[
\mathrm{{Proj}_{SO(3)}}(k\mathbf{Y}_i^{ori}) 
= k\mathbf{Y}_i^{ori}(k^2\mathbf{I})^{-1/2} 
= \mathbf{Y}_i^{ori}.
\]
Therefore, in this isotropic case the projection acts as an identity mapping, 
and we have $\mathbf{T}_i = \mathbf{Y}_i^{ori}$.
Consequently, $\mathbf{T}_i$ follows the same Matrix Fisher distribution as $\mathbf{Y}_i^{ori}$,
i.e.,
\[
\mathbf{T}_i \sim \mathcal{M}(\cdot; \mathbf{O\Lambda}_i).
\]
This result preserves the conjugate-like update form while ensuring 
the flow remains strictly on the $\mathrm{SO}(3)$ manifold and numerically stable. Empirically, we found that using a distribution with squared concentration parameter, $\mathbf{T}_i \sim\mathcal{M}(\cdot; \mathbf{O\Lambda}_i^{2})$, works better in practice. We therefore set
\begin{equation}
p_F(\mathbf{T}_i \mid \mathbf{O}, \mathcal{P}; t_i) = \mathcal{M}(\mathbf{T}_i; \mathbf{O} \boldsymbol{\Lambda}^2_i).
\end{equation}
This surrogate preserves the mean direction and increases concentration consistently with the linear scaling, while ensuring that the network always receives a rotation matrix as input. 
We stress that $\mathcal{M}(\cdot; \mathbf{O\Lambda}_i^{2})$ here is a \emph{computationally convenient approximation} (not the exact marginalized posterior); empirically, this projection-refit step is efficient and effective for our tasks.
\end{proof}

\section{Detailed BFNs for Residue Centroids and Residue Types} \label{app:centroids}

Here, we give detailed information on key components within the Bayesian flow framework.
\subsection{Euclidean BFN for Centroids}
Recall that the centroids (i.e., $C_\alpha$ positions) of $N$ residues in a peptide, denoted as $\mathbf{X}$, can be characterized by Gaussian distribution, i.e., $\mathbf{X}=\{\mathbf{x}^{(i)}\}_{i=1}^N \sim p_I(\cdot\mid \boldsymbol{\theta}^{pos}) = \mathcal{N}(\mathbf{X}\mid \boldsymbol{\mu}^{pos}, (\rho^{pos})^{-1}\mathbf{I})$, where $\boldsymbol{\theta}^{pos}\stackrel{\mathrm{def}}{=} \{\boldsymbol{\mu}^{pos}, \rho^{pos}\} $. 

We define the sender distribution as Gaussian distribution $p_S(\mathbf{y}^{pos}\mid \mathbf{X}; \alpha^{pos}) = \mathcal{N}(\mathbf{y}^{pos}\mid \mathbf{X}; \alpha^{pos})$. The receiver distribution can be characterized as $p_R(\mathbf{y}^{pos}\mid \boldsymbol{\theta}^{pos}, \mathcal{P};t_i) = \mathcal{N}(\mathbf{y}^{pos}\mid \hat{\mathbf{X}};(\alpha^{pos})^{-1}I)$, where $\hat{\mathbf{X}}$ is the output of a neural network $\Psi^{pos}(\boldsymbol{\theta}^{pos},\mathcal{P};t_i)$. The Bayesian flow distribution $p_F(\boldsymbol{\theta}^{pos}\mid \mathbf{X}, \mathcal{P}; t_i)=\mathcal{N}(\boldsymbol{\mu}\mid \gamma(t_i)\mathbf{X}, \gamma(t_i)(1-\gamma(t_i))I)$, where $\gamma(t_i)=1-(\sigma_1^{pos})^{2t_i}$. The neural network $\Psi^{pos}$ is trained by minimizing the KL divergence between the sender and receiver distributions:
\begin{align}
&L_n^{pos}(\mathbf{O}, \mathcal{P}) = n \, \mathbb{E}_{i \sim U(1, n),\, \boldsymbol{\theta}_{i-1}^{pos} \sim p_F} \, D_{\mathrm{KL}}(p_S \, \| \, p_R) \nonumber \\
&= \frac{n}{2} \, \mathbb{E}_{i \sim U(1, n),\, \boldsymbol{\theta}_{i-1}^{pos} \sim p_F} \, \alpha_i^{pos} \| \mathbf{X} - \hat{\mathbf{X}} \|^2.
\end{align}

The final centroids $\mathbf{X}_\text{final}$ can be sampled by initializing from simple priors and evolving their parameters through discrete time steps $i \in \{0, 1, \dots, N\}$. Specifically, centroids are initialized with parameters $\boldsymbol{\theta}_0^{pos} = \{\mathbf{0}, I\}$. The Bayesian update trajectories are as follows:
\begin{equation}
\boldsymbol{\theta}_0^{pos} 
\xrightarrow{\Psi^{pos}} \hat{\mathbf{X}} 
\xrightarrow{P_F} \boldsymbol{\theta}_1^{pos}
\rightarrow \cdots \rightarrow
\boldsymbol{\theta}_N^{pos}
\xrightarrow{\Psi^{pos}} \mathbf{X}_\text{final}. \\
\end{equation}

\subsection{Categorical BFN for Residue Types}

The amino acid types in a peptide can be denoted as $C = \{c^{(i)}\}_{i=1}^{D}$ where $c^{(i)} \in \{ c \in \mathbb{Z}^+ \mid 1 \leq c \leq 20 \} $. Here, we utilize a categorical distribution $\boldsymbol{\theta}^{type}=(\boldsymbol{\theta}^{(1)},\cdots,\boldsymbol{\theta}^{(D)})$ with $\boldsymbol{\theta}^{(d)} = \big(\theta^{(d)}_1, \cdots, \theta^{(d)}_K\big) \in \Delta^{K-1}$, where $\theta^{(d)}_k$ is the probability assigned to class k for $d$-th residue type. Then the input distribution gives $p_I(C\mid\boldsymbol{\theta}^{type})=\prod_{d=1}^D \theta^{(d)}_{c^{(d)}}$.

The sender distribution gives $p_S(y^{type}\mid C; \alpha^{type})=\mathcal{N}(\mathbf{y}^{type}\mid \alpha^{type}(K\mathbf{e}_c-1); \alpha^{type}KI)$, where $\boldsymbol{e}_c\stackrel{\mathrm{def}}{=}(\boldsymbol{e}_{c^{(1)}},\cdots,\boldsymbol{e}_{c^{(D)}}) \in \mathbb{R}^{KD}$ and $\boldsymbol{e}_{c^{(d)}} \in \mathbb{R}^K$ is the projection from the class index $c^{(d)}$ to the length $K$ one-hot vector. 

The receiver distribution gives:
\begingroup\scriptsize
\begin{align} 
    &p_R(\mathbf{y}^{type} \mid \boldsymbol{\theta}^{type}, \mathcal{P}; t, \alpha^{type})  \\
    &= \prod_{d=1}^D \sum_{k=1}^K p_O^{(d)}(k \mid \boldsymbol{\theta}^{type},\mathcal{P};t_{i-1}))\,
\mathcal{N}\big(\alpha^{type} (K e_k - 1), \alpha^{type} K I \big),\nonumber
\end{align}
\endgroup
where $p_O^{(d)}(k \mid \cdot)=\big(\mathrm{softmax}(\Psi^{(d)}(\cdot))\big)_k$. The Bayesian flow distribution is 
\begin{align}
&p_F(\boldsymbol{\theta}^{type} \mid C, \mathcal{P}; t_i)  \\
&= \mathbb{E}_{\mathcal{N}(\mathbf{y}^{type} \mid \beta(t_i)(K e_c - 1), \beta(t_i)K I)} \big[ \delta(\theta - \mathrm{softmax}(\mathbf{y}^{type})) \big], \nonumber
\end{align}
which can be sampled by drawing $\mathbf{y}^{type}$ from  $\mathcal{N}(\beta(t_i)(K e_c - 1), \beta(t_i)K I)$  and then setting $\boldsymbol{\theta} = \mathrm{softmax}(\mathbf{y})$.

The loss function gives:
\begingroup\scriptsize
\begin{align}
&L_n^{type}(\mathbf{x}) = n \, 
\mathbb{E}_{\substack{
i \sim U(1,n),\, 
p_F(\boldsymbol{\theta}^{type} \mid C, \mathcal{P}; t_{i-1}), \\
\mathcal{N}(\mathbf{y}^{type} \mid 
\alpha^{type}_i (K e_c - 1), \alpha^{type}_i K I)
}} \\
&\bigg[
\ln \mathcal{N}(\mathbf{y}^{type}
\mid\cdot) - \sum_{d=1}^D \ln 
\bigg( \sum_{k=1}^K p_O^{(d)}(k \mid \boldsymbol{\theta}^{type},\mathcal{P};t_{i-1}) \,
\mathcal{N}\big(y^{(d)}\mid\cdot\big) \bigg)
\bigg].\nonumber 
\end{align}
\endgroup

The final residue types $\mathbf{C}_\text{final}$ can be sampled by initializing from simple priors and evolving their parameters through discrete time steps $i \in \{0, 1, \dots, N\}$. Specifically, residue types are initialized with $\boldsymbol{\theta}_0^{type} = \tfrac{1}{K}\mathbf{1}$. Both follow similar Bayesian update trajectories:
\begin{equation}
\boldsymbol{\theta}_0^{type} 
\xrightarrow{\Psi^{type}} \hat{\mathbf{C}} 
\xrightarrow{P_F} \boldsymbol{\theta}_1^{type}
\rightarrow \cdots \rightarrow
\boldsymbol{\theta}_N^{type}
\xrightarrow{\Psi^{type}} \mathbf{C}_\text{final}.
\end{equation}

\section{Experiment Setup}\label{sec:train_test}

\begin{algorithm}[htb]
\caption{Simulate Bayesian Flow for Gaussian Mixture Distributions}
\label{alg:simu_GMM}
\begin{algorithmic}[1]
\State \textbf{Input:} $\chi, \boldsymbol{\theta}^{ang}_0 = \big\{ \mu^k_0, \rho^k_0, \pi^k_0 \big\}_{k=1}^K,$
\Statex $\text{angle likelihood precision}\{\alpha_i\}_{i=1}^N$
\State \textbf{Output:} Temporal Gaussian mixture parameters $\big\{\boldsymbol{\theta}_i^{ang}\big\}_{i=1}^N$
\For{$i = 1$ to $N$}
    \State Sample  $y_i \sim \mathcal{N}(\chi, \alpha^{-1}_i)$
    \For{each $k$}
    \State Compute marginal likelihood variance:
    \Statex \quad \quad \quad \quad $(\sigma^k_i)^2 \gets \alpha_i^{-1} + (\rho_{i-1}^k)^{-1}$
    \State Compute likelihood weight for component $k$:
    \Statex \quad \quad \quad \quad $p_i^k \gets \frac{1}{\sqrt{2\pi (\sigma_i^k)^2}}
\exp\left[-\frac{(y_i - \mu_{i-1}^k)^2}{2(\sigma_i^k)^2}\right]$
    \State Compute unnormalized weights: 
    \Statex \quad \quad \quad \quad $\pi^k_i \gets \pi^k_{i-1} \cdot p^k_i$
    \EndFor
    \For{each $k$}
        \State Compute posterior parameters:
        \Statex \quad \quad \quad \quad $\pi^k_i \gets \pi^k_i / \sum_{k=1}^K \pi^k_i$
        \Statex \quad \quad \quad \quad$\rho_i^k \gets \rho_{i-1}^k + \alpha_i$
        \Statex \quad \quad \quad \quad$\mu^k_i \gets \left( \rho^k_{i-1} \cdot \mu^k_{i-1} + \alpha_i \cdot y_i \right) / \rho^k_i$
    \EndFor
    \State $\boldsymbol{\theta}_i^{ang} \gets \big\{ \mu^k_i, \rho^k_i, \pi^k_i \big\}$
\EndFor
\State \textbf{Return} $\big\{\boldsymbol{\theta}_i^{ang}\big\}_{i=1}^N$
\end{algorithmic}
\end{algorithm}

\subsection{Gaussian Mixture-based BFN with Numerical Simulations}

Starting from the initial Gaussian mixture parameters $\theta_0^{ang} = \{\mu_0^k, \rho_0^k, \pi_0^k\}_{k=1}^K$ and a given likelihood precision $\alpha$, the Algorithm~\ref{alg:simu_GMM} iteratively propagates the mixture parameters over $N$ time steps. At each step, an observation $y_i$ is sampled from $\mathcal{N}(\chi, \alpha_i^{-1})$, and for each mixture component $k$, we compute the marginal likelihood variance $(\sigma_i^k)^2$, evaluate the likelihood weight $p_i^k$ of the observation, and update the unnormalized and normalized mixture weights $\pi_i^k$. Using these updated weights, the posterior component parameters $(\mu_i^k, \rho_i^k, \pi_i^k)$ are then recalculated, yielding the temporal sequence of parameters $\{\theta_i^{ang}\}_{i=1}^N$. To ensure comprehensive coverage of the torsional space, we compute the Gaussian mixture trajectories for $\chi$ angles across the full domain $[0, 2\pi]$ at a resolution of $\pi/180$ radians. This simulation-based approach bypasses the need for closed-form marginalization, providing a practical approximation of the Bayesian flow distribution $p_F^k(\boldsymbol{\theta}^{ang}\mid\cdot)$.

\subsection{Noise Scheduler for Matrix Fisher-based BFN}

Following~\cite{graves2023bayesian}, we enforce the expected entropy of the input distribution
\begin{equation} \label{eq:matrix_entropy}
    \mathbb{E}_{p_F(\boldsymbol{T}_i\mid \mathbf{O\Lambda}^2_i; t_i)} 
H\big(p_I(\mathbf{O} \mid \boldsymbol{\theta}^{ori})\big)
\end{equation}
to decrease linearly with respect to $t_i$. Here, we cannot derive an analytical expression of this term, but empirically we found $\boldsymbol{\Lambda}_i = \frac{10}{\exp(2)-1}\cdot(\exp(2t_i)-1) I$ a good approximation.

\subsection{Training and Inference Pseudocodes}
Training and inference procedures are given in Algorithm~\ref{alg:pep_bfn_loss} and Algorithm~\ref{alg:sampling} respectively. Note that during inference, we adopt the noise reduced sampling strategy proposed by MolCARFT~\cite{qu2024molcraft} for residue centroids, types, and rotations (line 15 in Algorithm~\ref{alg:sampling}). For updating residue side-chain angle parameters, we utilize the native Bayesian update function (line 4-14 in Algorithm~\ref{alg:sampling}).

\begin{algorithm}[H]
\caption{Discrete-Time Loss for Training PepBFN}
\label{alg:pep_bfn_loss}
\begin{algorithmic}[1]
\State \textbf{Input:} Peptide sample $\mathcal{G} = \{\mathbf{X}, \mathbf{O}, \mathbf{C}, \boldsymbol{\chi}\}$, target $\mathcal{P}$, time horizon $n$, loss weights $\lambda_1$--$\lambda_4$
\State \textbf{Output:} Total loss $\mathcal{L}_n$
\State $i \sim U(1,n)$, $t_i \gets \frac{i-1}{n}$
\State Sample parameters from Bayesian flow distributions:

\Statex $\boldsymbol{\theta}^{pos}_{i-1} \sim p_F^{pos}(\cdot \mid \mathbf{X}, \mathcal{P}; t_i)$
\Statex $\boldsymbol{T}_{i-1} \sim p_F^{ori}(\cdot \mid \mathbf{O}, \mathcal{P}; t_i)$
\Statex $\boldsymbol{\theta}^{type}_{i-1} \sim p_F^{type}(\cdot \mid \mathbf{C}, \mathcal{P}; t_i)$
\Statex $\boldsymbol{\theta}^{ang}_{i-1} \sim p_F^{ang}(\cdot \mid \boldsymbol{\chi}, \mathcal{P}; t_i)$

\State Network predictions from sampled flow parameters:
\Statex $\hat{\mathbf{X}}, \hat{\mathbf{O}}, \hat{\mathbf{C}}, \hat{\boldsymbol{\chi}} \gets \Psi(\boldsymbol{\theta}^{pos}_{i-1}, \boldsymbol{T}_{i-1}, \boldsymbol{\theta}^{type}_{i-1}, \boldsymbol{\theta}^{ang}_{i-1}, \mathcal{P}, t_i)$

\State Compute KL objectives for each component:
\Statex $L_n^{pos} \gets \frac{n}{2} \cdot \alpha^{pos}_i \cdot \|\mathbf{X} - \hat{\mathbf{X}}\|^2$
\Statex $L_n^{ang} \gets \frac{n}{2} \cdot \alpha^{ang}_i \cdot \|\boldsymbol{\chi} - \hat{\boldsymbol{\chi}}\|^2$
\Statex $L_n^{type} \gets n \cdot\text{KL}\left(p_S^{type} \| p_R^{type} \right)$
\Statex $L_n^{ori} \gets n \cdot \lambda_i a(\lambda_i)\cdot (3 - \mathrm{tr}(\hat{\mathbf{O}}^\top \mathbf{O}))$

\State \Return $\mathcal{L}_n = \lambda_1 L_n^{pos} + \lambda_2 L_n^{ori} + \lambda_3 L_n^{type} + \lambda_4 L_n^{ang}$
\end{algorithmic}
\end{algorithm}

\begin{algorithm}[htb]
\caption{Inference Procedure of PepBFN}
\label{alg:sampling}
\begin{algorithmic}[1]
\State \textbf{Input:} Initial priors $\boldsymbol{\theta}_0^{pos}, \boldsymbol{T}_0, \boldsymbol{\theta}_0^{type}, \boldsymbol{\theta}_0^{ang}$, context $\mathcal{P}$, number of steps $N$, $\text{angle likelihood precision}\{\alpha_i\}_{i=1}^N$
\For{$i=1$ to $N$}
    \State Predict current variables:
\Statex \quad \quad $\hat{\mathbf{X}}, \hat{\mathbf{O}}, \hat{\mathbf{C}}, \hat{\boldsymbol{\chi}} \gets \Psi(\boldsymbol{\theta}^{pos}_{i-1}, \boldsymbol{T}_{i-1}, \boldsymbol{\theta}^{type}_{i-1}, \boldsymbol{\theta}^{ang}_{i-1}, \mathcal{P}, t_i)$
    
    \State Sample from sender distributions:
    \Statex \quad \quad $y^{ang} \sim \mathcal{N}(\hat{\boldsymbol{\chi}}, (\alpha_i)^{-1})$
    \State Gaussian mixture-based Bayesian update for angles:
    \Statex \quad \quad $\big\{ \mu^k_i, \rho^k_i, \pi^k_i \big\}_{k=1}^K \gets\boldsymbol{\theta}^{ang}_i$
    \For{each $k$}
    \State Update marginal likelihood variance:
    \Statex \quad \quad \quad \quad $(\sigma^k_i)^2 \gets \alpha_i^{-1} + (\rho_{i}^k)^{-1}$
    \State Update likelihood weight for component $k$:
    \Statex \quad \quad \quad \quad $p_i^k = \frac{1}{\sqrt{2\pi (\sigma_i^k)^2}}
\exp\left[-\frac{(y^{ang} - \mu_{i}^k)^2}{2(\sigma_i^k)^2}\right]$
    \State Update unnormalized weights: 
    \Statex \quad \quad \quad \quad $\pi^k_i \gets \pi^k_{i} \cdot p^k_i$
    \EndFor
    \For{each $k$}
        \State Update posterior parameters:
        \Statex \quad \quad \quad \quad $\pi^k_i \gets \pi^k_i / \sum_{k=1}^K \pi^k_i$
        \Statex \quad \quad \quad \quad$\rho_i^k \gets \rho_{i}^k + \alpha_i$
        \Statex \quad \quad \quad \quad$\mu^k_i \gets \left( \rho^k_{i} \cdot \mu^k_{i} + \alpha_i \cdot y^{ang} \right) / \rho^k_i$
    \EndFor
    \State $\boldsymbol{\theta}_i^{ang} \gets \big\{ \mu^k_i, \rho^k_i, \pi^k_i \big\}$
    
    \State Bayesian update for centroids, orientations, and types using Bayesian flow distribution $p_F$:
    \Statex \quad \quad $\boldsymbol{\theta}^{pos}_{i} \sim p_F^{pos}(\cdot \mid \hat{\mathbf{X}}, \mathcal{P}; t_i)$
    \Statex \quad \quad $\boldsymbol{T}_{i} \sim p_F^{ori}(\cdot \mid \hat{\mathbf{O}}, \mathcal{P}; t_i)$
    \Statex \quad \quad $\boldsymbol{\theta}^{type}_{i} \sim p_F^{type}(\cdot \mid \hat{\mathbf{C}}, \mathcal{P}; t_i)$
\EndFor
\State \textbf{Return:} $\mathcal{G} = \Psi(\boldsymbol{\theta}_N^{pos}, \boldsymbol{T}_N, \boldsymbol{\theta}_N^{type},\boldsymbol{\theta}_N^{ang},\mathcal{P},t_i)$
\end{algorithmic}
\end{algorithm}

\subsection{Hyperparameters}

The model was trained using the Adam optimizer with a learning rate of 5e-4 on 8 RTX 4090 with a batch size of $10$ per GPU. We set $\lambda_1=1, \lambda_2=0.1, \lambda_3=1, \text{and} \ \lambda_4=1$, which represent the weights of translation, rotation, torsion, and sequence losses, respectively. The learning rate scheduling used the plateau strategy with a decay factor of 0.6, patience of 10 epochs, and minimum learning rate of 1e-6. Exponential moving average (EMA) was applied with a decay rate of 0.999.
We set $\rho_1^{pos}=1/0.03, \rho_1^{ang}=5, \lambda_1^{rot}=10, \text{and } \beta_1^{type}=1.2$. We trained PepBFN with 1000 discrete time steps, while using only 100 steps during generation for efficient sampling. The neural network $\Psi$ follows the architecture of PepFlow, with a key difference: our angle embedding is three times larger, as we concatenate the weighted means of $K = 3$ Gaussian components. This design allows the network to better capture the multimodal nature of side-chain torsion angles.

\begin{figure}[htbp]
    \centering
    \begin{subfigure}[b]{0.95\linewidth}
        \centering
        \includegraphics[width=\linewidth]{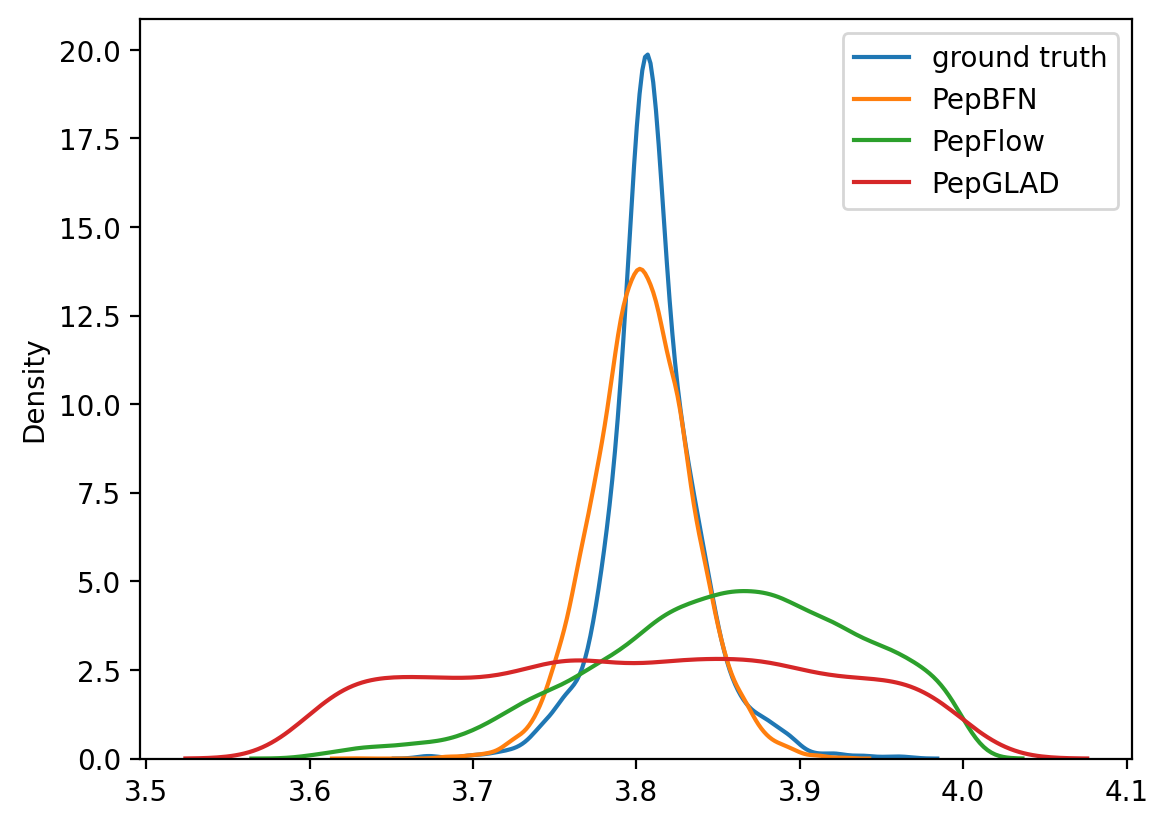}
        \caption{}
        \label{fig:sub1}
    \end{subfigure}
    \vskip 0.3cm
    \begin{subfigure}[b]{0.95\linewidth}
        \centering
        \includegraphics[width=\linewidth]{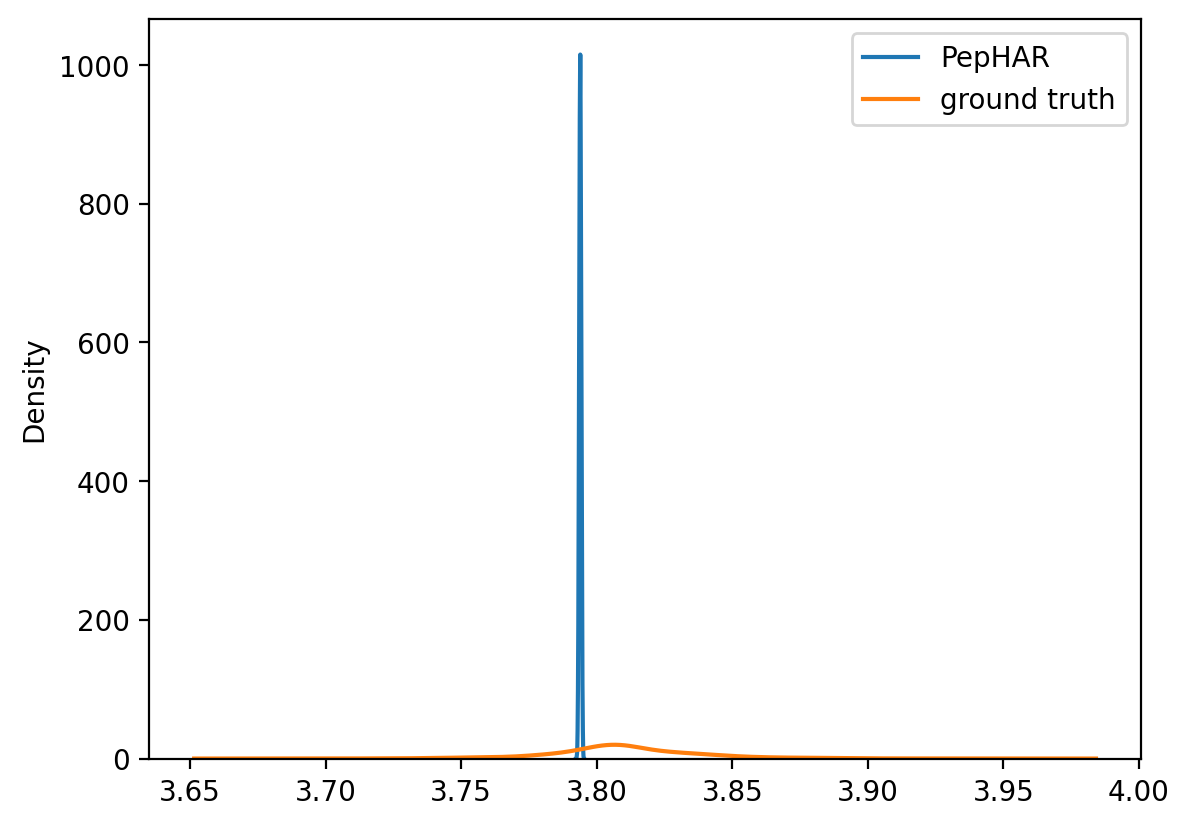}
        \caption{}
        \label{fig:sub2}
    \end{subfigure}
    \caption{(a, b) Distribution of distances between C$_\alpha$ atoms of amino acids. We compare the ground truth  with predictions from PepBFN, PepFlow, PepGLAD, and PepHAR, showing that PepBFN more accurately captures the native distance distribution.}
    \label{fig:ca_dist}
\end{figure}

\section{More Evaluation Results}

\subsection{Detailed $\boldsymbol{\chi}$ Angle MAE}

\begin{table}[H]
\centering
\renewcommand{\arraystretch}{1.15}
\begin{tabular}{llcccc}
\toprule
Type & $\chi$ & SCWRL4 & Rosetta & DLPacker & PepBFN \\
\midrule
LEU & 1 & 16.87 & 26.22 & 25.67 &  18.82\\
    & 2 & 24.84 & 41.55 & 42.51 &  19.21\\
\midrule
LYS & 1 & 45.31 & 50.43 & 54.45 &  34.04\\
    & 2 & 34.76 & 29.23 & 65.36 &  31.06\\
    & 3 & 20.88 & 30.35 & 47.66 &  25.17\\
    & 4 & 46.41 & 45.79 & 62.18 &  48.36\\
\midrule
MET & 1 & 27.12 & 49.00 & 25.92 &  8.03\\
    & 2 & 62.99 & 60.54 & 33.79 &  10.59\\
    & 3 & 61.00 & 83.63 & 65.07 &  54.10\\
\midrule
PHE & 1 & 14.33 & 30.86 & 23.91 &  4.38\\
    & 2 & 98.51 & 110.39 & 43.41 &  4.05\\
\midrule
SER & 1 & 62.66 & 48.56 & 43.89 &  3.98\\
\midrule
THR & 1 & 36.33 & 53.45 & 30.73 &  5.24\\
\midrule
TRP & 1 & 18.89 & 18.64 & 11.58 &  11.03\\
    & 2 & 27.73 & 31.44 & 26.37 &  8.71\\
\midrule
TYR & 1 & 21.04 & 27.99 & 35.46 &  13.66\\
    & 2 & 112.29 & 42.96 & 48.44 &  9.46\\
\midrule
VAL & 1 & 26.46 & 25.73 & 20.73 &  3.14\\
\midrule
CYS & 1 & 10.16 & 96.40 & 18.45 &  3.05\\
\midrule
ARG & 1 & 35.50 & 51.48 & 34.67 &  12.20\\
    & 2 & 43.07 & 41.25 & 39.48 &  15.98\\
    & 3 & 61.28 & 54.22 & 48.77 &  21.65\\
    & 4 & 59.14 & 71.71 & 65.48 &  60.52\\
\midrule
ASN & 1 & 29.22 & 35.57 & 12.09 &  4.38\\
    & 2 & 33.50 & 33.86 & 31.54 &  4.13\\
\midrule
ASP & 1 & 37.75 & 47.06 & 26.56 &  5.30\\
    & 2 & 76.96 & 78.44 & 84.31 &  4.17\\
\midrule
GLN & 1 & 48.98 & 63.41 & 52.91 &  11.59\\
    & 2 & 56.98 & 66.90 & 66.23 &  13.16\\
    & 3 & 53.08 & 71.95 & 70.21 &  49.55\\
\midrule
GLU & 1 & 55.72 & 69.74 & 63.63 &  18.51\\
    & 2 & 34.69 & 32.36 & 30.77 &  13.48\\
    & 3 & 60.28 & 63.67 & 69.88 &  37.39\\
\midrule
HIS & 1 & 12.48 & 15.02 & 20.99 &  3.51\\
    & 2 & 30.70 & 37.40 & 28.97 &  2.65\\
\midrule
ILE & 1 & 19.77 & 19.84 & 21.53 &  5.60\\
    & 2 & 38.75 & 49.52 & 43.69 &  6.57\\
\midrule
PRO & 1 & 12.86 & 12.75 & 10.95 &  2.83\\
    & 2 & 18.97 & 18.30 & 16.61 &  5.07\\
\bottomrule
\end{tabular}
\caption{Mean absolute error of different types of residues in side-chain prediction task.}
\label{tab:chi_angle}
\end{table}

Table~\ref{tab:chi_angle} summarizes the mean side-chain dihedral angle errors for all amino acids, comparing our method PepBFN with SCWRL4, Rosetta, and DLPacker. PepBFN achieves consistently lower or comparable errors across the majority of residues and $\boldsymbol{\chi}$ angles. In particular, PepBFN shows notable improvements for residues with fewer $\boldsymbol{\chi}$ angles, such as SER, THR, and VAL. For residues with more $\boldsymbol{\chi}$ angles, such as LYS, MET, and ARG, PepBFN also maintains moderate improved accuracy. This result aligns with the conclusion in DiffPack~\cite{zhang2023diffpack}, as simultaneously generating all $\boldsymbol{\chi}$ angles can lead to accumulated errors, whereas autoregressive generation mitigates this issue by modeling dependencies sequentially. Overall, these results indicate that PepBFN is highly effective in side-chain conformation prediction, achieving state-of-the-art performance across a wide range of amino acid types.

\subsection{Distance Distribution between $C_\alpha$ Atoms}

Fig.~\ref{fig:ca_dist} compares the distributions of $C_\alpha~–~C_\alpha$ distances under various methods against the ground truth. In panel (a), the ground truth exhibits a sharp peak near 3.8 \AA~ with a small but nonzero width, reflecting the natural geometric variability. PepBFN closely matches this behavior, capturing both the central tendency and a realistic spread, whereas PepFlow produces a broader, shifted distribution with heavier tails, indicating degraded distance precision and less concentrated modeling. PepGLAD yields an even more distorted and flattened distribution, deviating substantially from the native geometry. In panel (b), PepHAR collapses to an almost delta-like spike at 3.8 \AA, suggesting overconfidence and a lack of the intrinsic variability present in real structures. Overall, these results underscore that PepBFN achieves the best trade-off between geometric fidelity and plausible diversity in C$_\alpha$ distances, while the other methods either underfit (PepFlow, PepGLAD) or overfit/collapse (PepHAR).

\begin{figure}[htb]
    \centering
    \begin{subfigure}[b]{0.95\linewidth}
        \centering
        \includegraphics[width=\linewidth]{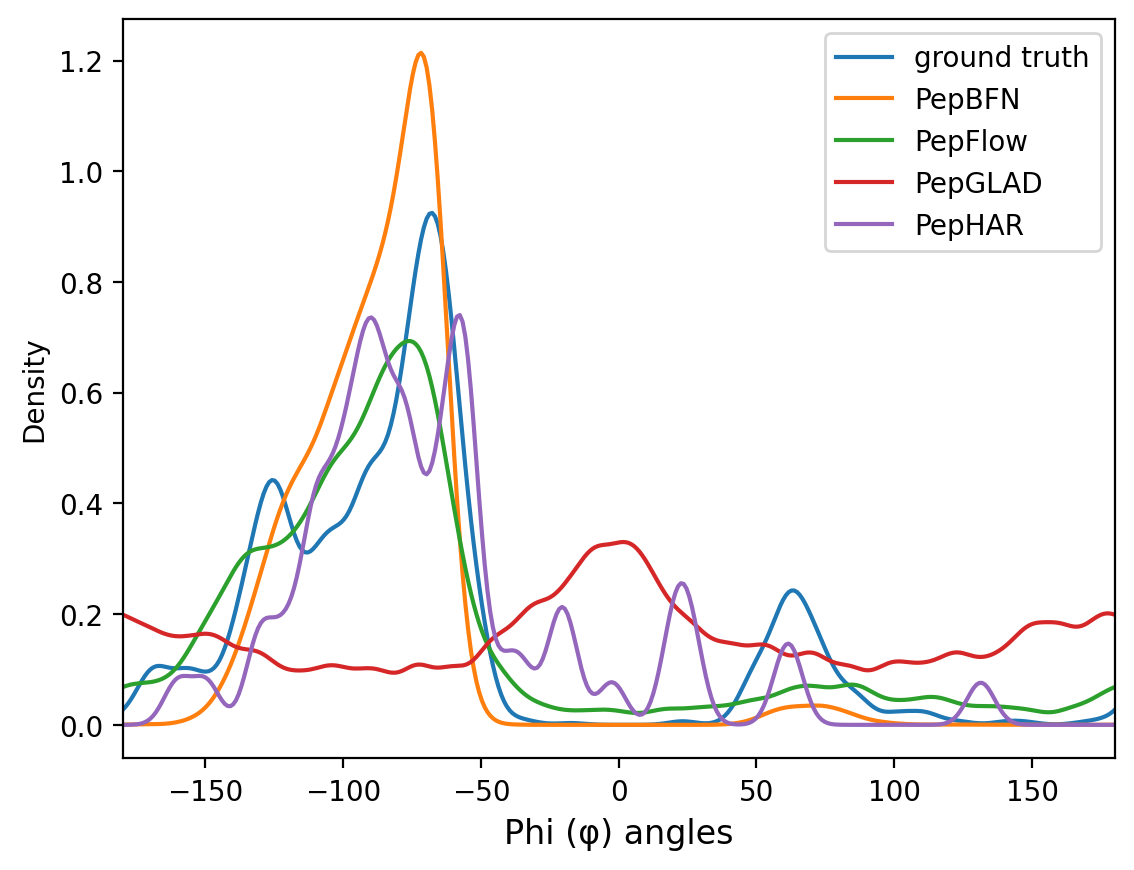}
        \caption{}
        \label{fig:phi}
    \end{subfigure}
    \vskip 0.3cm
    \begin{subfigure}[b]{0.95\linewidth}
        \centering
        \includegraphics[width=\linewidth]{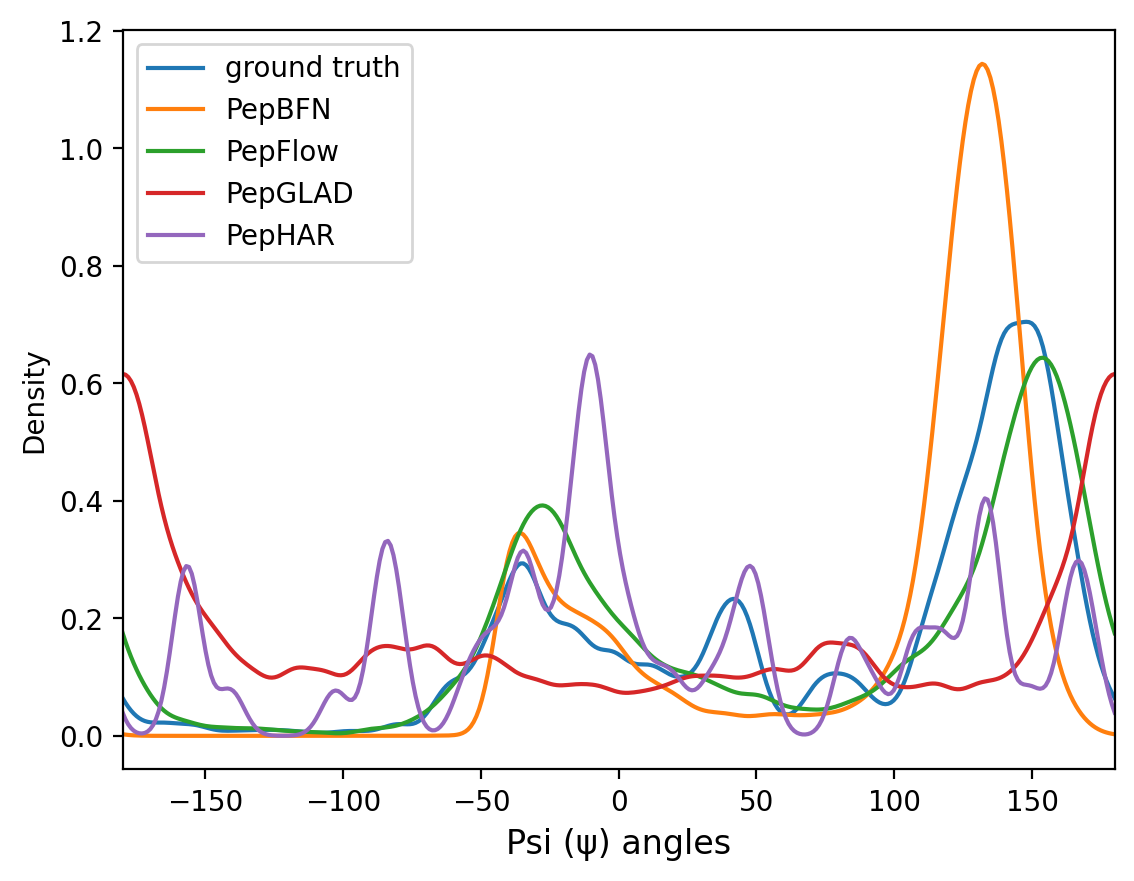}
        \caption{}
        \label{fig:psi}
    \end{subfigure}
    \caption{(a, b) Distribution of $\phi$ and $\psi$ angles, respectively. We compare the ground truth  with predictions from PepBFN, PepFlow, PepGLAD, and PepHAR.}
    \label{fig:phi_psi_dist}
\end{figure}

\subsection{Distributions of the $\phi$ and $\psi$ Angles}

Fig.~\ref{fig:phi_psi_dist} shows the empirical distributions of backbone dihedral angles \(\phi\) (panel a) and \(\psi\) (panel b) across different methods, with the ground truth as reference. PepBFN most faithfully reproduces the underlying multimodal structure of both angle distributions: it captures the dominant modes with high sharpness, indicating strong precision while introducing only modest bias in peak location and width. PepFlow recovers some major modes but with reduced contrast and less accurate relative heights, suggesting a tendency to underfit the conditional geometry. PepGLAD produces comparatively flattened and oversmoothed profiles, failing to resolve several distinct conformational states and thereby losing important structural detail. PepHAR, in contrast, displays spurious and overly concentrated peaks (notably in \(\psi\)) that do not align well with the ground truth, reflecting instability or collapse in its angular modeling. Overall, these results in the appendix underscore that PepBFN achieves the best trade-off between capturing true multimodality and maintaining sharpness, whereas the other methods either blur meaningful structure or introduce artifacts.

\subsection{Distributions of the Secondary Structure}

\begin{table}[H]
    \centering
\renewcommand{\arraystretch}{1.5}
    \begin{tabular}{cccc}
    \toprule
    Methods     &  Helix (\%) & Sheet (\%) & Coil (\%) \\
    \midrule
     test set    &  8.9 & 13.7 & 77.4 \\
     \hline
     PepHAR    &  4.1 & 0.8 & 95.1 \\
     \hline
     PepFlow    &  0.1 & 0 & 99.9 \\
     \hline
     PepBFN    &  5.3 & 1.0 & 93.7 \\
     \bottomrule
    \end{tabular}
    \caption{Empirical percentage of different secondary structures (helix, sheet, coil) for the test set and four generative models.}
    \label{tab:sscc}
\end{table}
Table~\ref{tab:sscc} shows empirical secondary structure distributions on the test set for the native structures and various generative models. All models predict significantly less helix and sheet than the ground truth, resulting in outputs that are overly dominated by coil. Recovering native secondary structure in short peptides is intrinsically challenging because peptides lack the extensive long-range and contextual interactions that stabilize helices and sheets in full proteins, making their conformations highly flexible and multimodal. Our future work with Bayesian Flow Networks will focus on explicitly incorporating secondary structure information into the generative process—e.g., via multi-task losses that jointly predict backbone dihedrals and target secondary structure labels, hierarchical conditioning where coarse-grained helix/sheet priors guide fine-grained angle generation, and augmenting the context with learned global sequence–structure embeddings to supply missing long-range cues. We also plan to explore combining BFNs with physics-aware refinement, such as local energy minimization, and transfer learning from larger, well-folded proteins to better regularize peptide sampling, aiming to reduce collapse toward coil and recover more balanced helix/sheet content.

\begin{figure}[htb]
    \centering
    \begin{subfigure}[b]{0.95\linewidth}
        \centering
        \includegraphics[width=\linewidth]{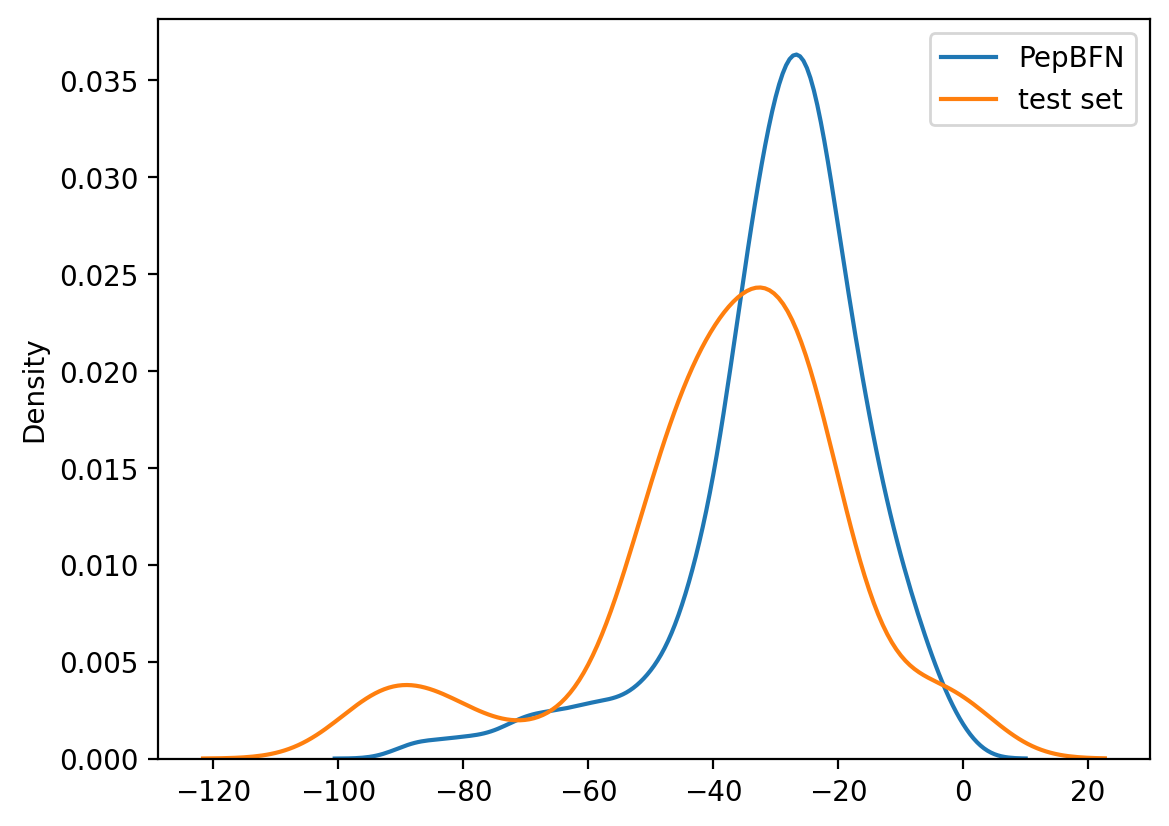}
        \caption{}
        \label{fig:aff}
    \end{subfigure}
    \vskip 0.3cm
    \begin{subfigure}[b]{0.95\linewidth}
        \centering
        \includegraphics[width=\linewidth]{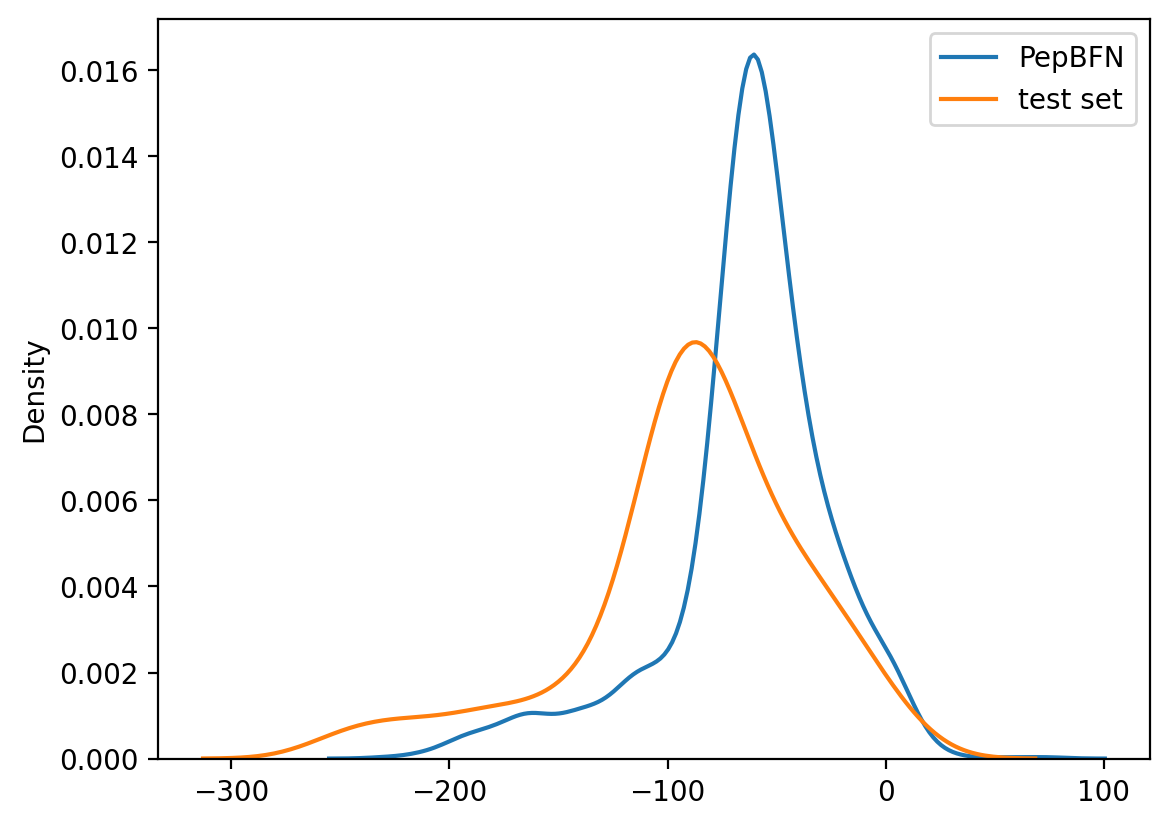}
        \caption{}
        \label{fig:stab}
    \end{subfigure}
    \caption{(a) Distributions of binding affinities. (b) Distributions of binding stabilities.}
    \label{fig:aff_stab}
\end{figure}

\subsection{Detailed Energy-related Distribution}

Fig.~\ref{fig:aff_stab} compares the learned PepBFN distributions to the empirical test-set distributions for two energy-related metrics, where lower values are better. Binding affinity measures the thermodynamic favorability of peptide–receptor association, while binding stability reflects the robustness of the bound state under realistic perturbations. PepBFN produces concentrated energy distributions whose modes are aligned well with those of the test set, indicating that the model captures the central tendency of both affinity and stability.

\subsection{Inference Time}

\renewcommand{\arraystretch}{2}
\begin{table}[h]\small
    \centering
    \begin{tabular}{ccc}
    \toprule
    Method     &  peptides per second  & CUDA memory per peptide\\
    \midrule
    PepGLAD    &  0.8  & $\sim$200MB \\
    \hline
    PepFLow    &  1.0  & $\sim$20MB \\
    \hline
    PepBFN    &  2.0  & $\sim$20MB \\
    \bottomrule
    \end{tabular}
    \caption{Inference speed and memory cost.}
    \label{tab:inference_benchmark}
    \vspace{-1em}
\end{table}

As shown in Table~\ref{tab:inference_benchmark}, PepBFN achieves the highest throughput (2 peptides/sec) with a low memory footprint (20MB per peptide). Generally speaking, compared to flow matching models, BFN converges in $\sim50\%$ fewer steps and $2\times$ speedup, with diffusion models being the slowest.

\section{More Visualizations}

\begin{figure*}
    \centering
    \includegraphics[width=0.85\linewidth]{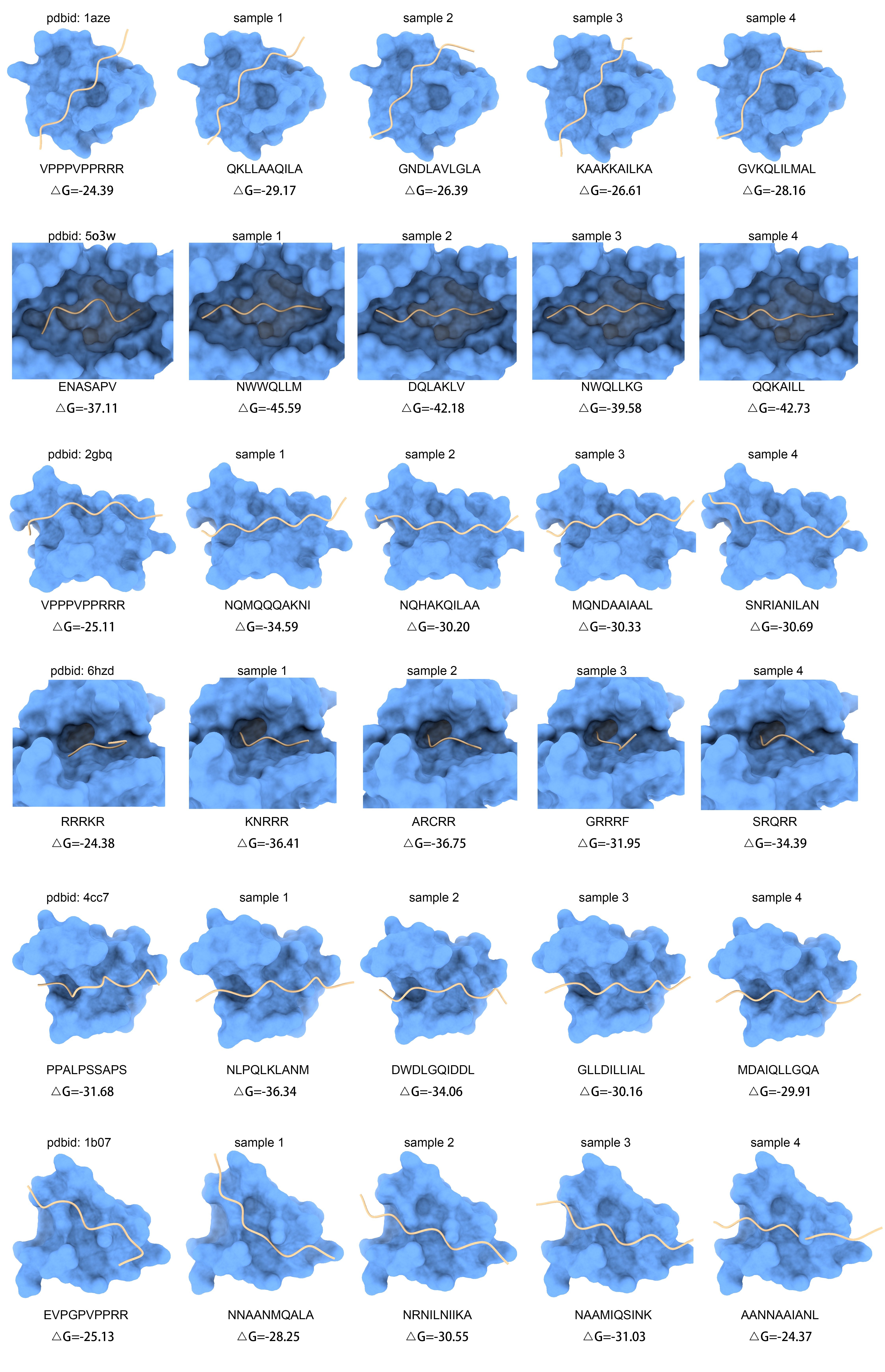}
    \caption{Additional generated peptides by PepBFN.}
    \label{fig:more_vis_1}
\end{figure*}

\begin{figure*}
    \centering
    \includegraphics[width=0.85\linewidth]{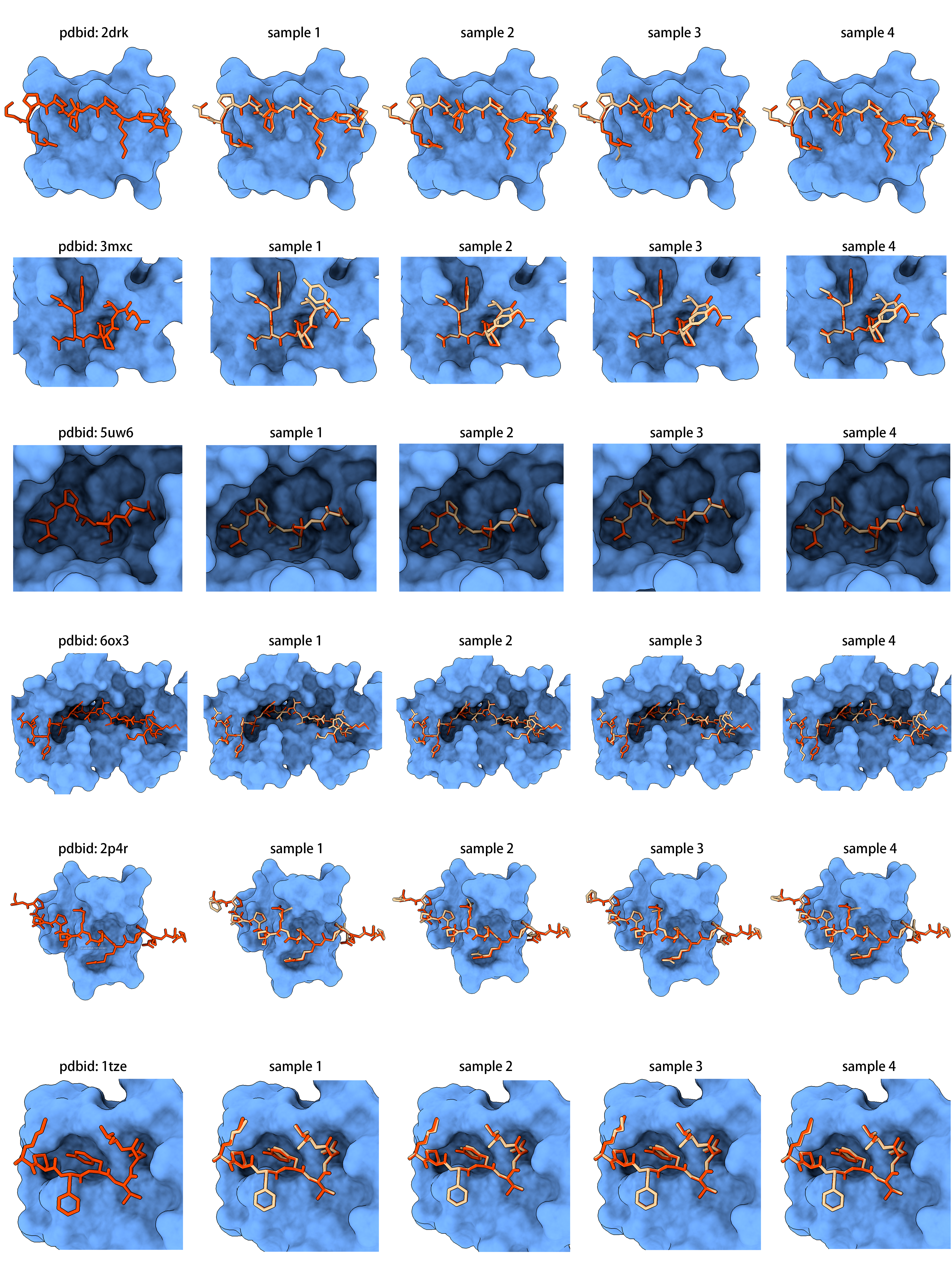}
    \caption{Overlay of the ground-truth (red) and PepBFN-predicted (yellow) peptide conformations within the binding pocket, with the protein surface rendered in blue, illustrating the close agreement between predicted and actual side-chain angles.}
    \label{fig:more_vis_2}
\end{figure*}

As illustrated in Fig.~\ref{fig:more_vis_1}, our method effectively designs diverse peptides with improved binding affinities compared to the known peptide. For the side-chain prediction task, as shown in Fig.~\ref{fig:more_vis_2}, our approach accurately recovers side-chain torsion angles. These results demonstrate the versatility and effectiveness of our method for a broad range of peptide design-related tasks.

\section{Broader Applications}

Due to its modular structure, PepBFN is not limited to full-atom peptide design, but can be readily extended to a range of protein and peptide modeling tasks. The centroid and orientation modules (translation and rotation) enable backbone-related applications such as loop modeling and de novo scaffold generation. The sequence module supports inverse folding and sequence recovery given backbone structures. The torsion module is well suited for side-chain packing, where accurate modeling of local rotameric states is essential. Moreover, the combination of translation, rotation, and torsion modules facilitates flexible peptide–protein docking, enabling precise interfacial positioning and conformational refinement. These capabilities make PepBFN a versatile tool for designing not only generic peptides, but also specialized biomolecules such as enzyme active sites, antibody-binding loops, thereby addressing diverse therapeutic and biotechnological needs.

\end{document}